\pdfoutput=1
\documentclass[sigconf]{acmart}

\AtBeginDocument{%
  \providecommand\BibTeX{{%
    \normalfont B\kern-0.5em{\scshape i\kern-0.25em b}\kern-0.8em\TeX}}}

\renewcommand\footnotetextcopyrightpermission[1]{}

\setcopyright{none} 

\usepackage{amsfonts,amsmath}
\usepackage{mathtools}
\usepackage{xcolor}
\usepackage{xspace}
\usepackage{hyperref}
\usepackage{multirow}
\usepackage[ruled,vlined]{algorithm2e}
\usepackage{enumitem}
\usepackage{booktabs}
\usepackage{subcaption}
\usepackage{color, colortbl}

\definecolor{Gray}{gray}{0.9}
\definecolor{CRed}{RGB}{255 204 204}
\definecolor{CGreen}{RGB}{102 255 102}
\definecolor{CBlue}{RGB}{153 204 255}
\usepackage{bm}
\newcommand{\adversary}{$\mathit{Adv}$\xspace}
\newcommand{\victim}{$\mathit{v}$\xspace}
\newcommand{\detector}{\texorpdfstring{AD$^3$\xspace}\xspace}
\newcommand{\uaps}{UAP-S\xspace}
\newcommand{\uapo}{UAP-O\xspace}
\newcommand{\osfw}{OSFW\xspace}
\newcommand{\osfwu}{OSFW(U)\xspace}
\newcommand{\vf}{VF\xspace}

\newcommand{\vect}[1]{\boldsymbol{#1}}
\newcommand{\s}{\ensuremath{\vect{s}}\xspace}
\newcommand{\qval}{\ensuremath{Q(\vect{s},a)}\xspace}
\newcommand{\vval}{\ensuremath{V(\vect{s})}\xspace}

\newcommand{\newtext}[1]{{#1}}
\definecolor{color1}{HTML}{1F77B4}
\definecolor{color2}{HTML}{FF7F0e}
\definecolor{color3}{HTML}{2CA02C}
\definecolor{color4}{HTML}{D62728}
\definecolor{color5}{HTML}{9467BD}

\newcommand{\removetext}[1]{}

\DeclareMathOperator*{\argmax}{argmax}

\usepackage[maxfloats=30,morefloats=12]{morefloats}

\begin{document}

\title{Real-time Adversarial Perturbations against Deep Reinforcement Learning Policies: Attacks and Defenses}

\author{Buse G.  A.  Tekgul}
\affiliation{%
\institution{Aalto University}
\city{Espoo}
\country{Finland}}
\email{batlitekgul@acm.org}

\author{Shelly Wang}
\affiliation{%
\institution{University of Waterloo}
\city{Waterloo}
\country{Canada}}
\email{shelly.wang@uwaterloo.ca}

\author{Samuel Marchal}
\affiliation{%
\institution{WithSecure Corporation \& Aalto University }
\city{Espoo}
\country{Finland}}
\email{samuel.marchal@aalto.fi}

\author{N. Asokan}
\affiliation{%
\institution{University of Waterloo \& Aalto University}
\city{Waterloo}
\country{Canada}}
\email{asokan@acm.org}

\renewcommand{\shortauthors}{Tekgul,  et al.}

\begin{abstract}
      Deep reinforcement learning (DRL) is vulnerable to adversarial perturbations. Adversaries can mislead the policies of DRL agents by perturbing the state of the environment observed by the agents. 
      Existing attacks are feasible in principle, but face challenges in practice, either by being too slow to fool DRL policies in real time or by modifying past observations stored in the agent's memory. We show that Universal Adversarial Perturbations (UAP), independent of the individual inputs to which they are applied, can fool DRL policies effectively and in \emph{real time}. We introduce three attack variants leveraging UAP. Via an extensive evaluation using three Atari 2600 games, we show that our attacks are effective, as they fully degrade the performance of three different DRL agents (up to 100\%, even when the $l_\infty$ bound on the perturbation is as small as 0.01). It is faster than the frame rate (60 Hz) of image capture and considerably faster than prior attacks ($\approx 1.8$ms). Our attack technique is also efficient, incurring an online computational cost of $\approx 0.027$ms. Using two tasks involving robotic movement, we confirm that our results generalize to complex DRL tasks. Furthermore, we demonstrate that the effectiveness of known defenses diminishes against universal perturbations. We introduce an effective technique that detects all known adversarial perturbations against DRL policies, including all universal perturbations presented in this paper.\footnote{The code reproducing our work is available in \url{https://github.com/ssg-research/ad3-action-distribution-divergence-detector}.}
\end{abstract}




\maketitle

\pagestyle{plain}

\section{Introduction}
\label{sec:Introduction}

\setcounter{footnote}{0}

Machine learning models are vulnerable to \emph{adversarial examples}: maliciously crafted inputs generated by adding small perturbations to the original input to force a model into generating wrong predictions~\cite{goodfellow2014explaining,szegedy2014intriguing}. Prior work~\cite{huang2017adversarial,kos2017delving,lin2017tactics} has also shown that adversarial examples can fool deep reinforcement learning (DRL) agents using deep neural networks (DNNs) to approximate their decision-making strategy.
If this vulnerability is exploited in safety-critical DRL applications such as robotic surgery and autonomous driving, the impact can be disastrous.

A DRL agent can partially or fully observe the \emph{state} of the environment by capturing complex, high-dimensional observations. For example, a DRL agent playing an Atari 2600 game observes pixels from each image frame of the game to construct states by combining a number of observations. DRL agents use the current state as an input to their \emph{policy} that outputs an optimal action for that state.
Consequently, adversaries can modify the environment to mislead the agent's policy. 
Various state-of-the-art attack methods assume \emph{white-box} knowledge, where adversaries have
access to the parameters of the agent's policy model and the reinforcement learning algorithm. 
In \emph{untargeted} attacks, the adversary aims to fool the agent's policy so that the agent 1) cannot complete its task or 2) finishes its task with unacceptably poor performance.
Prior work has shown that white-box attacks can successfully destroy agents' performance using one-step gradient-based approaches~\cite{goodfellow2014explaining}, optimization-based methods~\cite{carlini2017towards}, or adversarial saliency maps~\cite{papernot2016limitations}. Previous work has also proposed different attack strategies where the adversary generates the perturbation for 1) each state~\cite{behzadan2017vulnerability,huang2017adversarial}, 2) \emph{critical states} where the agent prefers one action with high confidence~\cite{kos2017delving,lin2017tactics,sun2020stealthy}, or 3) periodically, at every $N^{th}$ state ~\cite{kos2017delving}.
\begin{figure*}[t]
	\Description[<short description>]{<long description>}
    \centering
    \includegraphics[width=0.90\textwidth]{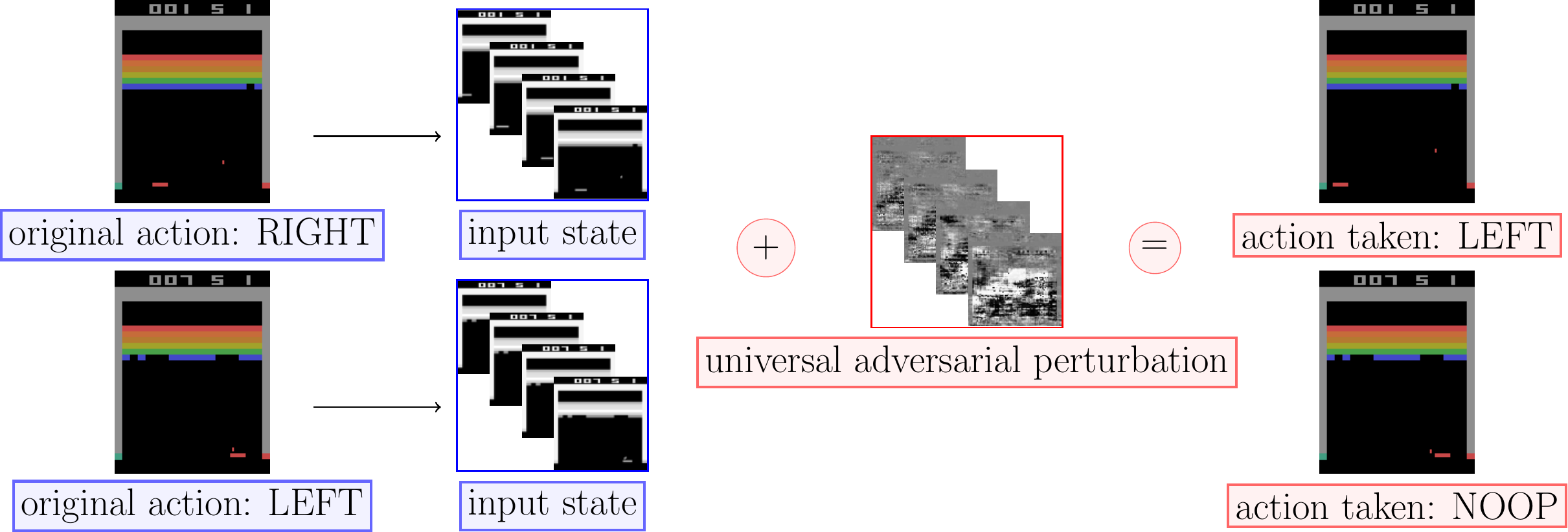}
    \caption{
   In the Breakout game, when a universal adversarial perturbation is added into different states, it causes the policy of a victim DQN agent to take sub-optimal actions instead of optimal actions, i.e., fail to catch the ball and lose the game.}\label{fig:introfig}
\end{figure*}

Although prior white-box attacks using adversarial perturbations are effective in principle, they are not realistic in practice. First, some attack strategies are based on computing the perturbation by solving an optimization problem~\cite{lin2017tactics}. This is computationally expensive, even if it is done for every $N^{th}$ state. DRL agents must respond to new states very quickly to carry out the task effectively \emph{on-the-fly}. Therefore, attacks that take longer than the average time between two consecutive observations 
are too slow to be realized in real-time. Second, in realistic scenarios, the adversary cannot have full control over the environment. However, iterative attacks~\cite{lin2017tactics,xiao2019characterizing}
require querying agents with multiple perturbed versions of the current state and resetting the environment to find the optimal perturbation. 
Therefore, iterative attacks cannot be applied in real-life scenarios, such as autonomous agents interacting with a dynamic environment. 
Finally, the aforementioned state-of-the-art attacks require seeing all observations to generate and apply perturbations to the state containing multiple observations. However, the agent can store clean observations that are part of the current state in its memory before the adversary can generate perturbations that need to be applied to \emph{all of those observations}.

\textbf{Contributions:} We propose an effective, \emph{real-time} attack strategy to fool DRL policies by computing state-agnostic \emph{universal} perturbations \emph{offline}. Once this perturbation is generated, it can be added into any state to force the victim agent to choose sub-optimal actions.(Figure~\ref{fig:introfig})
Similarly to 
previous work~\cite{goodfellow2014explaining,lin2017tactics,xiao2019characterizing}, we focus on untargeted attacks in a white-box setting. 
Our contributions are as follows: 
\begin{enumerate}[topsep=2pt,leftmargin=*,nolistsep]
	\item We design two new real-time white-box attacks, \uaps and \uapo, using Universal Adversarial Perturbation (UAP)~\cite{moosavi2017universal} to generate state-agnostic adversarial examples. 
	We also design a third real-time attack, \osfwu, by extending Xiao et al.'s~\cite{xiao2019characterizing} attack so that it generates a universal perturbation once, applicable to \emph{any subsequent episode}
	(Section~\ref{sec:Methodology}). An empirical evaluation of these three attacks using three different DRL agents playing three different Atari 2600 games (Breakout, Freeway, and Pong) demonstrates that our attacks are comparable to prior work in their effectiveness (100\% drop in return), while being \emph{significantly faster} (0.027 ms on average, compared to 1.8 ms) and \emph{less visible} 
	than prior adversarial 
	perturbations~\cite{huang2017adversarial,xiao2019characterizing} (Section~\ref{ssec:attackperformance}).
	Using two additional tasks that involve the MuJoCo robotics simulator~\cite{todorov2012mujoco}, which requires continuous control, we show that our results generalize to more complex tasks (Section~\ref{ssec:performancecontinuous}).
	
	\item 
	We demonstrate the limitations of prior defenses.
	We show that agents trained with the state-of-the-art robust policy regularization technique~\cite{zhang2020robust} exhibit reduced effectiveness against adversarial perturbations at higher perturbation bounds $(\geq 0.05)$. In some tasks (Pong), universal perturbations completely destroy agents' performance (Table~\ref{tab:attack_summary}).
	Visual Foresight~\cite{lin2017detecting}, which is another defense method that can restore an agent's performance in the presence of prior adversarial perturbations~\cite{huang2017adversarial}, fails to do so when faced with universal perturbations (Section~\ref{subsec:defense-effectiveness}).
	\item We propose an efficient method, \detector, to detect adversarial perturbations. \detector can be combined with other defenses to provide stronger resistance for DRL agents against untargeted adversarial perturbations (Section~\ref{subsec:ad3}).
\end{enumerate}
\section{Background and Related Work}
\label{sec:Background}

\subsection{Deep Reinforcement Learning}
\label{ssec:drl}

\subsubsection{Reinforcement learning}
Reinforcement learning involves settings where an agent continuously interacts with a non-stationary environment to decide which action to take in response to a given state.
At time step $t$, the environment is characterized by its state $\vect{s} \in \mathcal{S}$ consisting of $N$ past observations $o$ pre-processed by some function $f_{pre}$, i.e., $\vect{s} = \{f_{pre}(o_{t-N+1}), \cdots, f_{pre}(o_{t})\} $. At each \s, the agent takes an action $a \in \mathcal{A}$, which moves the environment to the next $\vect{s{'}} \in \mathcal{S}$ and receives a reward $r$ from the environment. The agent uses this information to optimize its policy $\pi$, a probability distribution that maps states into actions~\cite{Sutton1998}. During training, the agent improves the estimate of a value function \vval or an action-value function \qval. \vval measures how valuable it is to be in a state \s by calculating the \emph{expected discounted return}: the discounted cumulative sum of future rewards while following a specific $\pi$. Similarly, \qval estimates the value of taking action $a$ in the current \s while following $\pi$. During evaluation, the optimized $\pi$ is used for decision making, and the \emph{performance} of the agent is measured by the return. 
In this work, we focus on \emph{episodic}~\cite{bellemare2013arcade} and \emph{finite-horizon}~\cite{tassa2018deepmind} tasks. In episodic tasks such as Atari games, each episode ends with a terminal state (e.g., winning/losing a game, arriving at a goal state), and the return for one episode is computed by the total score in single-player games (e.g., Breakout, Freeway) or the relative score when it is played against a computer (e.g., Pong). Finite-horizon tasks include continuous control, where (e.g., Humanoid, Hopper) the return is measured for a fixed length of the episode.

\subsubsection{DNN}
DNNs are parameterized functions $f(\vect{x}, \theta)$ consisting of neural network layers. For an input $\vect{x} \in \mathbb{R}^n$ with $n$ features, the parameter vector $\theta$ is optimized by training $f$ over a labeled training set. $f(\vect{x}, \theta)$ outputs a vector $\vect{y} \in \mathbb{R}^{m}$ with $m$ different classes. In classification problems, the predicted class is denoted as $\hat{f}(\vect{x}) = \operatorname*{argmax}_{m}f(\vect{x},\theta)$. For simplicity, we will use $f(\vect{x})$ to denote $f(\vect{x}, \theta)$. 

\subsubsection{DRL} 
DNNs are useful for approximating $\pi$ when $\mathcal{S}$ or $\mathcal{A}$ is too large, or largely unexplored. Deep Q Networks (DQN) is one of the well-known \emph{value-based} DRL algorithms~\cite{mnih2015human} that uses DNNs to approximate \qval. 
During training, DQN aims to find the optimal \qval, and defines the optimal $\pi$ implicitly using the optimal \qval. 
Despite its effectiveness, DQN cannot be used in continuous control tasks, where $a$ is a real-valued vector sampled from a range, instead of a finite set. Continuous control tasks require \emph{policy-based} DRL algorithms~\cite{mnih2016asynchronous,schulman2017proximal}. They use two different DNNs that usually share a number of lower layers to approximate both $\pi$ and \vval (or \qval),
and update $\pi$ directly. For example, in \emph{actor-critic} methods (A2C)~\cite{mnih2016asynchronous}, the critic estimates \vval or \qval for the current $\pi$, and the actor updates the parameter vector $\theta$ of $\pi$ by using \emph{advantage}, which refers to the critic's evaluation of the action decision using the estimated \vval or \qval, while the actor follows the current $\pi$. 
Proximal Policy Optimization (PPO)~\cite{schulman2017proximal} is another on-policy method that updates the current $\pi$ by ensuring that the updated $\pi$ is close to the old one. 

\subsection{Adversarial Examples}\label{sec:adversarialexamples}
An adversarial example $\vect{x}^{*}$ against a classifier $f$ is a deliberately modified version of an input $\vect{x}$ such that $\vect{x}$ and $\vect{x}^*$ are similar, but $\vect{x}^*$ is misclassified by $f$, i.e., $\hat{f}(\vect{x})\neq\hat{f}(\vect{x}^*)$. An untargeted adversarial example is found by solving%
\begin{gather}\label{eq:advexamples}
      \argmax_{\vect{x}^*}\ell(f(\vect{x}^*),\hat{f}(\vect{x}))\;
    \text{\small s.t.: } \lVert{\vect{x}^*-\vect{x}}\rVert_{p}  = \lVert{\vect{r}}\rVert_p \leq \epsilon,
\end{gather}
where $\epsilon$ is the $l_p$ norm bound and $\ell$ is the loss between $f(\vect{x}^*)$ and the predicted label $\hat{f}(\vect{x})$. In this work, we use the $l_{\infty}$ norm bound (i.e., any element in the perturbation must not exceed a specified threshold $\epsilon$) as in the state-of-the-art Fast Gradient Sign Method (FGSM)~\cite{goodfellow2014explaining}. FGSM calculates $\vect{r}$ by 
\begin{equation}
    \vect{r} = \epsilon \cdot \text{sign}({\nabla_{\vect{x}}\ell(f(\vect{x}),\hat{f}(\vect{x})})).
\end{equation}

Adversarial examples are usually computed for each $\vect{x}$. An alternative is to generate input-agnostic \emph{universal perturbations}. For instance, Moosavi et al. propose \emph{Universal Adversarial Perturbation} (UAP)~\cite{moosavi2017universal} that searches for a sufficiently small $\vect{r}$ that can be added to the \emph{arbitrary} inputs to yield adversarial examples against $f$. UAP iteratively computes a unique $\vect{r}$ that fools $f$ for almost all inputs $\vect{x}$ belonging to a training set $\mathcal{D}_{train}$. UAP utilizes DeepFool~\cite{moosavi2016deepfool} to update $\vect{r}$ at each iteration.
UAP aims to achieve the desired \emph{fooling rate} $\delta$: the proportion of successful adversarial examples against $f$ with respect to the total number of perturbed samples $|\mathcal{D}_{train}|$.

Following the first work~\cite{moosavi2017universal} introducing UAP, many different strategies~\cite{co2019procedural,hayes2018learning,mopuri2017fast,mopuri2018nag} have been proposed to generate universal adversarial perturbations to fool image classifiers. For example, Hayes et al.~\cite{hayes2018learning} and Mopuri et al.~\cite{mopuri2018nag} use generative models to compute universal adversarial perturbations. Mopuri et al.~\cite{mopuri2017fast} introduce the Fast Feature Fool algorithm, which does not require a training set to fool the classifier, but optimize a random noise to change activations of the DNN in different layers. 
Co et al.~\cite{co2019procedural} design black-box, untargeted universal adversarial perturbations using procedural noise functions.

\subsubsection{Adversarial Examples in DRL}\label{ssec:advDRL}
In discrete action spaces with finite actions, 
adversarial examples against $\pi$ are found by modifying Equation~\ref{eq:advexamples}: $\vect{x}$ is changed to \s at time step $t$, $f$ is replaced with \qval, and $\hat{Q}(\vect{s})$ refers to the decided action.
We also denote $Q(\vect{s}, a_m)$ as the state-action value of $m^{th}$ action at \s. In this setup, adversarial examples are computed to decrease \qval for the optimal action at \s, resulting in a sub-optimal decision. 

Since Huang et al.~\cite{huang2017adversarial} showed the vulnerability of DRL policies to adversarial perturbations, several untargeted attack methods~\cite{behzadan2017vulnerability,kos2017delving,lin2017tactics,sun2020stealthy,xiao2019characterizing} have been proposed that manipulate the environment. Recent work~\cite{baluja2018learning,carlini2017towards,hussenot2020copycat,lin2017tactics,tretschk2018sequential} has also developed targeted attacks, where the adversary's goal is to lure the victim agent into a specific state or to force the victim policy to follow a specific path. 
Most of these methods implement well-known adversarial example generation methods such as FGSM~\cite{huang2017adversarial,kos2017delving}, JSMA~\cite{behzadan2017vulnerability} and Carlini\&Wagner~\cite{lin2017tactics}.
Therefore, even though they effectively decrease the return of the agent, these methods cannot be implemented in real-time and have a temporal dependency: they need to compute a different $\vect{r}$ for every \s.
 
Similarly to our work, Xiao et al. propose (``obs-seq-fgsm-wb'', \osfw) to generate universal perturbations. \osfw computes a single $\vect{r}$ by applying FGSM to the averaged gradients over the $k$ states and adds $\vect{r}$ to the remaining states in the current episode. However, \osfw has limitations, such as the need to compute $\vect{r}$ for every new episode. Moreover, \osfw has to freeze the task to calculate $\vect{r}$, and its performance depends on the particular agent-environment interaction for each episode. Hussenot et al.~\cite{hussenot2020copycat} design targeted attacks that use different universal perturbations for each action to force the victim policy to follow the adversary's policy.

In addition to white-box attacks, multiple black-box attack methods based on finite-difference methods~\cite{xiao2019characterizing}, and proxy methods that approximate the victim policy~\cite{inkawhich2019snooping,zhao2020blackbox} were proposed, but they cannot be mounted in real time, as they require querying the agent multiple times. Recent work~\cite{gleave2019adversarial,wu2021adversarial} also shows that adversaries can fool DRL policies in multi-agent, competitive games by training an adversarial policy for the opponent agent that exploits vulnerabilities of the victim agent. These attacks rely on creating natural observations with adversarial effects, instead of manipulating the environment by adding adversarial perturbations. In this paper, we focus on single-player games, where adversaries can only modify the environment to fool the DRL policies.
\section{State- and Observation-Agnostic Perturbations}
\label{sec:Methodology}
\subsection{Adversary Model}
\label{sec:advModel}
The goal of the adversary \adversary is to degrade the performance of the victim DRL agent \victim by adding perturbations $\vect{r}$ to the state \s observed by \victim.
\adversary is successful when the attack:
\begin{enumerate}[nolistsep]
	\item is \emph{effective}, i.e., limits \victim to a low return,
	\item is \emph{efficient}, i.e., can be realized in real-time, and
	\item \emph{evades} known detection mechanisms.
\end{enumerate}

\adversary has a white-box access to \victim; therefore, it knows \victim's action value function $Q_{\mathit{v}}$, or the policy $\pi_{\mathit{v}}$ and the value function $V_{\mathit{v}}$, depending on the DRL algorithm used by \victim.
However, \adversary is constrained to using $\vect{r}$ with a small norm to evade possible detection, either by specific anomaly detection mechanisms or via human observation. We assume that \adversary cannot reset the environment or return to an earlier state. In other words, we rule out trivial attacks (e.g., swapping one video frame with another or changing observations with random noise) as ineffective because they can be easily detected. \newtext{We also assume that \adversary has only read-access to \victim's memory. Modifying the agent's inner workings is an assumption that is too strong in realistic adversarial settings because it forces \adversary to modify both the environment and \victim. If \adversary is able to modify or rewrite \victim's memory, then it does not need to compute adversarial perturbations and simply rewrites \victim's memory to destroy its performance.}

\subsection{Attack Design}\label{subsec:UAPAttack}

\subsubsection{Training data collection and sanitization}\label{ssec:trainingData}
\adversary collects a training set $\mathcal{D}_{train}$ by monitoring \victim's interaction with the environment for one episode, and saving each \s into $\mathcal{D}_{train}$. Simultaneously, \adversary clones $Q_{\mathit{v}}$ or $V_{\mathit{v}}$ into a \emph{proxy agent}, $adv$. 
Specifically, in value-based methods, \adversary copies the weights of $Q_{\mathit{v}}$ into $Q_{adv}$. In policy-based methods, \adversary copies the weights of the critic network into $V_{adv}$. In the latter case, \adversary can obtain $Q_{adv}(\s, a)$ by calculating $V_{adv}(\s)$ for each \emph{discrete} action $a \in \mathcal{A}$.

After collecting $\mathcal{D}_{train}$, \adversary sanitizes it by choosing only the \emph{critical states}. Following~\cite{lin2017tactics}, we define critical states as those that can have a significant influence on the course of the episode. We identify critical states using the relative action preference function
\begin{equation}\label{eqn:actionpreference}
\begin{aligned}
&\text{Var}_{a \in \mathcal{A}}\left[ \text{Softmax}(Q_{adv}(\s, a)) \right] \geq \beta \\
\beta &= 1/|\mathcal{D}_{train}| \; \sum_{\quad \mathclap { \s \in \mathcal{D}_{train}}} {\text{Var}_{a \in \mathcal{A}}\left[ \text{Softmax}(Q_{adv}(\s, a)) \right] },
\end{aligned}
\end{equation}
modified from~\cite{lin2017tactics}, where $\mathrm{Var}$ is the variance of the normalized $Q_{adv}(\s, a)$ values computed for $ \forall a \in \mathcal{A}$. 
This ensures that both \uaps and \uapo are optimized to fool $Q_{\mathit{v}}$ in critical states, and achieves the first attack criterion.

\subsubsection{Computation of perturbation}
For both \uaps and \uapo, we assume that $s \in \mathcal{D}_{train}$ and $ \mathcal{D}_{train} \subset \mathcal{S}$. \adversary searches for an optimal $\vect{r}$ that satisfies the constraints in Equation~\ref{eq:advexamples}, while achieving a high fooling rate $\delta$ on $\mathcal{D}_{train}$. 
For implementing \uaps and \uapo, we modify Universal Adversarial Perturbation~\cite{moosavi2017universal} (see Section~\ref{sec:adversarialexamples}). 
The goal of both \uaps and \uapo is to find a sufficiently small $\vect{r}$ such that $\hat{Q}(\vect{s}+\vect{r}) \neq \hat{Q}(\vect{s})$, leading \victim to choose sub-optimal actions. Algorithm~\ref{algo:uapso} summarizes the method for generating \uaps and \uapo.

\begin{algorithm}[t]\LinesNumbered
	\caption{Computation of \uaps and \uapo}\label{algo:uapso}
	\SetKwInOut{Input}{input}
	\SetKwInOut{Output}{output}
	\Input{\text{sanitized }$\mathcal{D}_{train}, Q_{adv}, \text{desired fooling rate } \delta_{th},$ \\
		$\text{max. number of iterations } it_{max}, \text{ perturbation constraint } \epsilon $}
	\Output{universal $\vect{r}$}
	Initialize $\vect{r} \gets 0, it\gets 0$\;
	\While{$\delta < \delta_{max}$ \textbf{ and }  $it < it_{max}$}{
		\For{$\vect{s} \in \mathcal{D}_{train}$}{
			\If{$\hat{Q}(\vect{s}+\vect{r}) = \hat{Q}(\vect{s})$ }{
				{Find the extra, minimal $\Delta\vect{r}$:}
				$\Delta \vect{r} \gets \text{argmin}_{\Delta \vect{r}} {\lVert \Delta \vect{r} \rVert}_{2} \text{ s.t.: }
				\hat{Q}(\vect{s}+\vect{r}+\Delta \vect{r}) \neq \hat{Q}(\vect{s})$\;
				$\vect{r} \gets \text{sign}(\text{min}(\text{abs}
				(\vect{r} + \Delta \vect{r}), \epsilon))$\;}
		}
		{Calculate $\delta$ with updated $\vect{r}$ on $\mathcal{D}_{train}$}\;
		$it \gets (it+1)$\;
	}
\end{algorithm}

In lines 5-6 of Algorithm~\ref{algo:uapso}, UAP-S and UAP-O utilize DeepFool to compute $\Delta \vect{r}$ by iteratively updating the perturbed 
$\vect{s}^{*}_i = \vect{s} + \vect{r} + \Delta \vect{r}_i$ until $\hat{Q}_{adv}$ outputs a wrong action (see Algorithm 2 in~\cite{moosavi2016deepfool}).
At each iteration $i$, DeepFool finds the closest hyperplane $\hat{l}(\vect{s}^{*}_{i})$ and $\Delta \vect{r}_i$ that projects $\vect{s}^{*}_{i}$ on the hyperplane. It recomputes $\Delta \vect{r}_i$ as

\begin{align}
	Q^{'}(\vect{s}^{*}_i, a_{\hat{l}}) & \gets Q_{adv}(\vect{s}^{*}_i, a_{\hat{l}}) - Q_{adv}(\vect{s}^{*}_i, a_m), \nonumber \\
	\vect{w}^{'}_{\hat{l}} &\gets \nabla Q_{adv}(\vect{s}^{*}_i, a_{\hat{l}})- \nabla Q_{adv}(\vect{s}^{*}_i, a_m), \nonumber \\
	\Delta \vect{r}_i &\gets 
	\frac{| Q^{'}(\vect{s}^{*}_i, a_{\hat{l}}) |}{ {\lVert {\vect{w}^{'}_{\hat{l}}} \rVert}^{2}_{2} }\vect{w}^{'}_{\hat{l}},
\end{align}

where $\nabla$ is the gradient of $Q_{adv}$ w.r.t. $\vect{s}_i$ and $Q_{adv}(\vect{s}^{*}_i, a_m)$ is the value of the $m$-th action chosen for the state $\vect{s} + \vect{r}$. 

\uaps computes a different perturbation for each observation $o_j$ in \s, i.e., $\vect{r}= \{r_{t-N+1}, \cdots, r_{t}\},\, r_j \neq r_k,  \, \forall j, k \in \{t-N+1, \cdots, t\},\, j \neq k$. In contrast, \uapo applies the same perturbation to all observations in \s. Therefore, it can be considered as an observation-agnostic, completely universal attack. \uapo aims to find a modified version $\tilde{\vect{r}}$ of $\vect{r}$ by solving 
\begin{gather}\label{eqn:uapoconstraints}
	\operatorname*{min}({\lVert \vect{r} - \tilde{\vect{r}} \rVert}^{2}_{2}) \\
	\begin{aligned}
		\text{\small s.t.: }
		\tilde{r}_j = \tilde{r}_k, \, \forall j,k \in \{t-N+1, \cdots, t\}\;
		\text{\small and } \lVert{\tilde{\vect{r}}}\rVert_{\infty} \leq \epsilon. \nonumber \\
	\end{aligned}
\end{gather}

In \uapo, we modify lines 5-6 of Algorithm~\ref{algo:uapso} to find $\Delta \vect{r}$. The closest $\Delta \tilde{\vect{r}}_i$ to $\Delta \vect{r}_i$ satisfying the conditions of Equation~\ref{eqn:uapoconstraints} is found by averaging $\vect{w}^{'}_{\hat{l}}$ over observations:

\begin{equation}\label{eqn:uapo}
\Delta \tilde{r}_{i_j}\gets  
\frac{| Q^{'}(\vect{s}^{*}_i, a_{\hat{l}}) |}{ N{\lVert {\vect{w}^{'}_{\hat{l}}} \rVert}^{2}_{2} }
\quad \smashoperator{\sum_{k=(t-N+1)}^{t}}\;\vect{w}^{'}_{\hat{l}_{k}},\, \forall j \in \{t-N+1, \cdots, t\}.
\end{equation}

In \uapo, DeepFool returns $\Delta \tilde{\vect{r}}_i =\Delta \tilde{r}_{i_j}$ as the optimal additional perturbation. 
\uapo adds the same $\tilde{r}_j$ to every $o_j$ in \s. If \s consists of only one observation, then \uaps will simply reduce to \uapo. The proof of Equation~\ref{eqn:uapo} can be found in Appendix~\ref{sec:proof}.

\subsubsection{Extending \osfw to \osfwu}

As explained in Section~\ref{ssec:advDRL}, \osfw calculates $\vect{r}$ by averaging gradients of $Q_{\mathit{v}}$ using the first k states in an episode and then adds $\vect{r}$ to the remaining states. This requires 1) generating a different $\vect{r}$ for each episode, and 2) suspending the task (e.g., freezing or delaying the environment) and \victim to perform backward propagation. Moreover, the effectiveness of \osfw varies when \victim behaves differently in individual episodes. We extend \osfw to a completely universal adversarial perturbation by using the proxy agent's DNN, and calculate averaged gradients with first k samples from the same, un-sanitized $\mathcal{D}_{train}$. The formula for calculating $\vect{r}$ in \osfwu is
\begin{equation}
\vect{r} = \epsilon \cdot \text{sign}(1/k
\sum_{i=0}^{i < k}
\nabla_{\vect{s}_i}  (- \log(Q_{adv}(\vect{s}_{i}, \hat{a})))
)
,
\end{equation}\label{eqn:oswfu}
where $\hat{a}$ denotes the action chosen and $\vect{s}_i \in \mathcal{D}_{train}$.

\subsection{Attacks in Continuous Control Settings}
In continuous control tasks, the optimal action is a real-valued array that is selected from a range. These tasks have complex environments that involve physical system control such as Humanoid robots with multi-joint dynamics~\cite{todorov2012mujoco}. Agents trained for continuous control tasks have no $Q$ that can be utilized for generating perturbations to decrease the value of the optimal action. Nevertheless, \adversary can find a perturbed state $\vect{s} + \vect{r}$ that has the worst \vval, so that $\pi_{\mathit{v}}$ might produce a sub-optimal action~\cite{zhang2020robust}. In continuous control, \osfw and \osfwu can be simply modified by changing $Q_{adv}$ with $V_{adv}$. However, in Algorithm~\ref{algo:uapso}, the lines 4-6 need to be adjusted to handle these tasks. \adversary can only use the copied network parameters of $V_{adv}$ to find $\vect{r}$. The corrected computation for \uaps and \uapo is given in Algorithm~\ref{algo:uapcontinuous}. In each iteration $it$, $\Delta \vect{r}$ is computed by DeepFool~\cite{moosavi2016deepfool} (lines 5 and 6).

\begin{algorithm}[t]\LinesNumbered
\caption{Computation of UAP in continuous control}\label{algo:uapcontinuous}
\SetKwInOut{Input}{input}
\SetKwInOut{Output}{output}
\Input{\text{sanitized }$\mathcal{D}_{train}, V_{adv}$, \text{hyper-parameter } $\alpha$ \\
$\text{max. number of iterations } it_{max}, \text{ pert. constraint } \epsilon $}
\Output{universal $\vect{r}$}
Initialize $\vect{r} \gets 0, it \gets 0$\;

\While{$\delta < \delta_{max}$ \textbf{ and }  $it < it_{max}$}{
\For{$\vect{s} \in \mathcal{D}_{train}$}{
\If{$V_{adv}(\vect{s}+\vect{r})+\alpha < V_{adv}(\vect{s})$ }{
{Find the extra, minimal $\Delta\vect{r}$:}\\
$\Delta \vect{r} \gets \text{argmin}_{\Delta \vect{r}} {\lVert \Delta \vect{r} \rVert}_{2}$ \\
\text{ s.t. }
{$V_{adv}(\vect{s}+\vect{r}+\Delta \vect{r}) +\alpha  < V_{adv}(\vect{s})$}\;
$\vect{r} \gets \text{sign}(\text{min}(\text{abs}(\vect{r} + \Delta \vect{r}), \epsilon))$\;}
}
{Calculate $\delta$ with updated $\vect{r}$ on $\mathcal{D}_{train}$}\;
$it \gets (it+1)$\;
}
\end{algorithm}

\section{Attack evaluation}
\label{sec:Evaluation}

\subsection{Experimental Setup}

We compared the effectiveness and efficiency of our attacks (\uaps, \uapo and \osfwu) with prior attacks (FGSM and \osfw) on discrete tasks using three Atari 2600 games (Pong, Breakout, Freeway) in the Arcade Learning Environment~\cite{bellemare2013arcade}. We further extended our experimental setup with the MuJoCo robotics simulator~\cite{todorov2012mujoco} and compared these attacks in continuous control tasks.

\begin{figure*}[t]
	\Description[<short description>]{<long description>}
	\centering
	\includegraphics[width=0.95\textwidth]{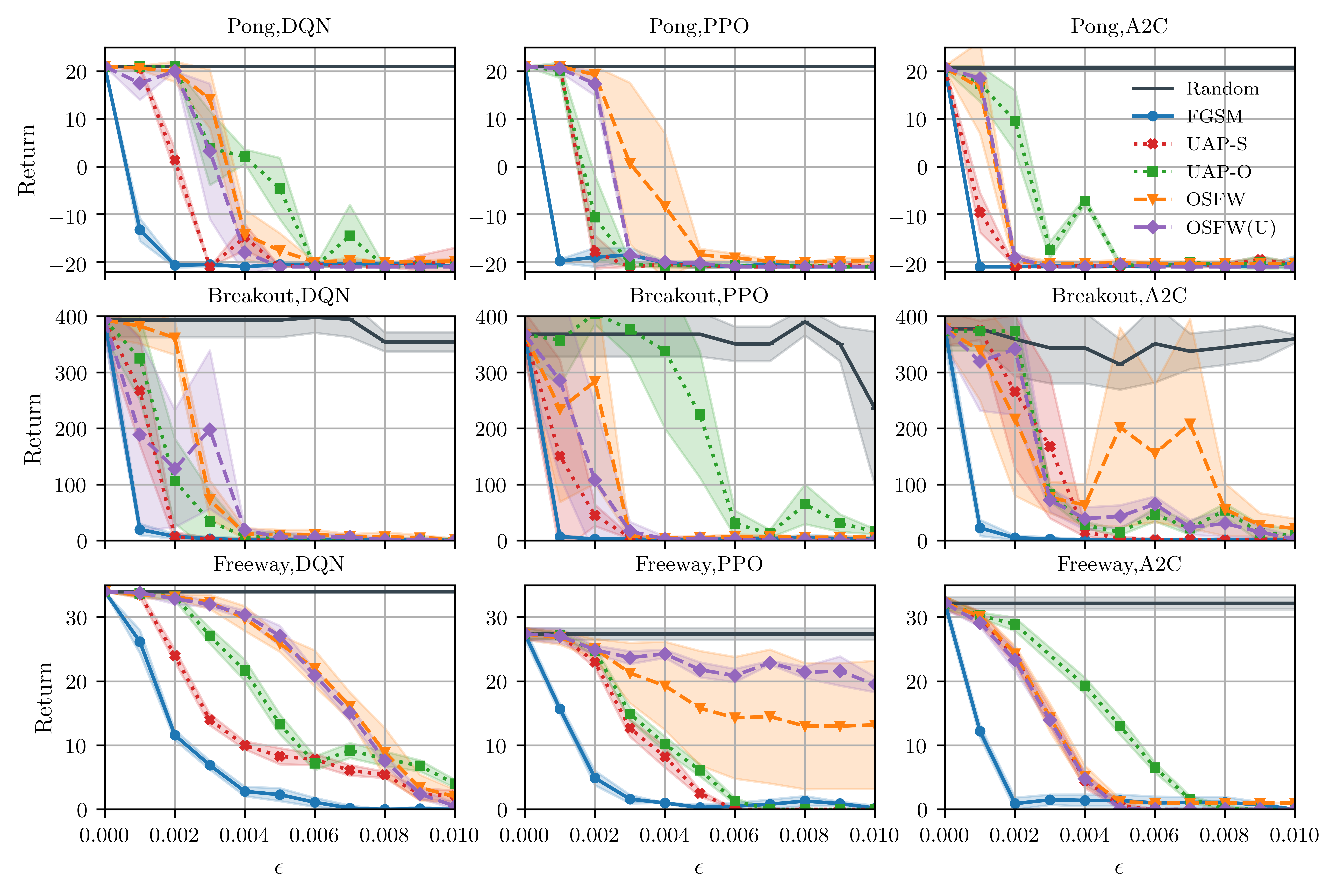}
	\caption{Comparison of attacks against DQN, PPO and A2C agents trained for Pong, Breakout and Freeway. The graph shows how the return (averaged over 10 games) changes with different $\epsilon$ values for six different attack strategies.}\label{fig:attackcomparison}
\end{figure*}

To provide an extensive evaluation, for every Atari game, we trained three agents each using a different DRL algorithm: value-based DQN~\cite{mnih2015human}, policy-based PPO~\cite{schulman2017proximal} and actor-critic method A2C~\cite{mnih2016asynchronous}. We used the same DNN architecture proposed in~\cite{mnih2015human} for approximating $Q_{\mathit{v}}$ in DQN and $V_{\mathit{v}}$ in other algorithms. 
To facilitate comparison, we used the same setup to implement all DRL policies as well as the different attacks: PyTorch (version 1.2.0), NumPy (version 1.18.1), Gym (a toolkit for developing reinforcement learning algorithms, version 0.15.7) and MuJoCo 1.5 libraries. All experiments were done on a computer with 2x12 core Intel(R) Xeon(R) CPUs (32GB RAM) and NVIDIA Quadro P5000 with 16GB memory.

Our implementations of DQN, PPO and A2C are based on OpenAI baselines\footnote{\url{https://github.com/DLR-RM/rl-baselines3-zoo}}, and our implementations achieve similar returns as in OpenAI Baselines. For pre-processing, we converted each RGB frame to gray-scale, re-sized those from $210\times160$ to $84\times84$, and normalized the pixel range from $[0, 255 ]$ to $[0, 1]$. We used the frame-skipping technique~\cite{mnih2015human}, where the victim \victim constructs $\vect{s}$ at every $N{th}$ observation, then selects and action and repeats it until the next state $\vect{s}'$. WE set $N=4$, so the input state size is $4 \times 84\times 84$. The frame rate of each game is 60 Hz by default~\cite{bellemare2013arcade}; thus, the time interval between two consecutive frames is $1/60=0.017$ seconds.

We implemented \uaps and \uapo by setting the desired $\delta_{max}$ to $95\%$, so that they stop searching for another $\vect{r}$ 
when $\delta \geq 95\%$. As baselines, we used random noise addition, FGSM~\cite{huang2017adversarial}, and \osfw~\cite{xiao2019characterizing}. We chose FGSM as the baseline since it is the fastest adversarial perturbation generation method from the previous work and is effective in degrading the performance of DRL agents. We measured attack effectiveness where $ 0 \leq \epsilon \leq 0.01$. We reported the average return over 10 episodes and used different seeds during training and evaluation.

\subsection{Attack Performance}\label{ssec:attackperformance}

\subsubsection{Performance degradation} Figure~\ref{fig:attackcomparison} compares \uaps, \uapo, and \osfwu with two baseline attacks and random noise addition. Random noise addition cannot cause a significant drop in \victim's performance, and FGSM is the most effective attack, reducing the return up to 100\% even with a very small $\epsilon$ value. 
\uaps is the second most effective attack in almost every setup, reducing the return by more than $50\%$ in all experiments when $\epsilon \geq 0.004$. All attacks completely destroy \victim's performance at $\epsilon = 0.01$, except the PPO agent playing Freeway. The effectiveness of \uapo, \osfw, and \osfwu is comparable in all setups. 
We also observe that the effectiveness of \osfw fluctuates heavily (Breakout-A2C) or has a high variance (Freeway-PPO). This phenomenon is the result of the different behaviors of \victim in individual episodes (Breakout-A2C), and \osfw's inability to collect enough knowledge (Freeway-PPO) to generalize $\vect{r}$ to the rest of the episode.

\subsubsection{Timing comparison} Table~\ref{tab:testtiming_comparison} shows the computational cost to generate $\vect{r}$, and the upper bound on the online computational cost to mount the attack in real-time. This upper bound is measured as $T_{max} = 1/\text{(frame rate)}-\text{(response time)}$, where the response time is the time spent feeding $\vect{s}$ forward through $\pi_{\mathit{v}}$ or $Q_{\mathit{v}}$ and executing the corresponding action. If the online cost of injecting $\vect{r}$ during deployment (and the online cost of generating $\vect{r}$ during deployment in the case of FGSM and \osfw) is greater than $T_{max}$, then \adversary must stop or delay the environment, which is infeasible in practice. Table~\ref{tab:testtiming_comparison} confirms that the online cost of \osfw is higher than that of all other attacks and $T_{max}$ due to its online perturbation generation approach. \osfw has to stop the environment to inject $\vect{r}$ into the current $\vect{s}$ or has to wait 
102 states on average $(\text{online cost}/(T_{max}\cdot N))$ to correctly inject $\vect{r}$, which can decrease the attack effectiveness. \uaps and \uapo have a higher offline cost than \osfwu, but the offline generation of $\vect{r}$ does not interfere with the task, as it does not require interrupting or pausing \victim. The online cost of FGSM, \uaps, \uapo and \osfwu is lower than $T_{max}$. 

\begin{table}[t]
	\centering
	\resizebox{1.0\columnwidth}{!}{
		\begin{tabular}{llcc}
			\hline
			\bf \multirow{2}{*}{Experiment} & \bf Attack & \bf Offline cost$\pm$ std & \bf Online cost $\pm$ std  \\ 
			& \bf method & \bf (seconds) & \bf (seconds) \\ \hline
			\multirow{5}{0.4\linewidth}{
				Pong, DQN, $T_{max} = 0.0163\pm10^{-6}$ \\ seconds
			} 
			& FGSM &  - &  $13\times10^{-4}\pm10^{-5}$\\
			& \osfw & - & \textcolor{red}{$5.3\pm0.1$} \\
			& \uaps & $36.4\pm21.1$ & $2.7\times10^{-5}\pm10^{-6}$ \\
			& \uapo & $138.3\pm25.1$& $2.7\times10^{-5}\pm10^{-6}$ \\
			& \osfwu & $5.3\pm0.1$ & $2.7\times10^{-5}(\pm10^{-6})$ \\ \hline
			\multirow{5}{0.4\linewidth}{
				Pong, PPO, \\ $T_{max} = 0.0157\pm10^{-5}$ \\ seconds
			} 
			& FGSM & - & $21\times10^{-4}\pm10^{-5}$ \\
			& \osfw & - & \textcolor{red}{ $7.02\pm0.6$} \\
			& \uaps & $41.9\pm16.7$ & $2.7\times10^{-5}\pm10^{-6}$ \\
			& \uapo & $138.3\pm25.1$& $2.7\times10^{-5}\pm10^{-6}$ \\
			& \osfwu & $7.02\pm0.6$ & $2.7\times10^{-5}\pm10^{-6}$ \\ \hline
			\multirow{5}{0.4\linewidth}{
				Pong, A2C \\ $T_{max} = 0.0157\pm10^{-5}$ \\ seconds
			} & FGSM &  - &   $21\times10^{-4}\pm10^{-5}$\\
			& \osfw & -  & \textcolor{red}{$7.2\pm1.1$}  \\
			& \uaps & $11.4\pm4.3$ & $2.7\times10^{-5}\pm10^{-6}$ \\
			& \uapo & $55.5\pm29.3$ & $2.7\times10^{-5}\pm10^{-6}$ \\
			& \osfwu & $7.2\pm1.1$ & $2.7\times10^{-5}\pm10^{-6}$ \\ \hline
		\end{tabular}}
			\caption{Offline and online cost of attacks 
				where victim agents are DQN, PPO, A2C trained for Pong and $\epsilon=0.01$. Attacks that cannot be implemented in real-time are in highlighted in red.}
			\label{tab:testtiming_comparison}
	\end{table}
    
\subsubsection{Difference in perturbation sizes} 
\begin{figure*}[t]
	\Description[<short description>]{<long description>}
	\centering
	\begin{subfigure}{.13\linewidth}
		\centering
		\includegraphics[width = \linewidth]{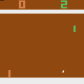}
		\caption*{Clean, RGB}
	\end{subfigure}%
	\hspace{1.0em}%
	\begin{subfigure}{.13\linewidth}
		\centering
		\includegraphics[width = \linewidth]{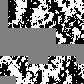}
		\caption*{FGSM}
	\end{subfigure}%
	\hspace{1.0em}%
	\begin{subfigure}{.13\linewidth}
		\centering
		\includegraphics[width = \linewidth]{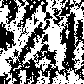}
		\caption*{\osfw}
	\end{subfigure}%
	\hspace{1.0em}%
	\begin{subfigure}{.13\linewidth}
		\centering
		\includegraphics[width = \linewidth]{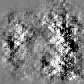}
		\caption*{\uaps}
	\end{subfigure}
	\hspace{1.0em}%
	\begin{subfigure}{.13\linewidth}
		\centering
		\includegraphics[width = \linewidth]{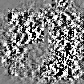}
		\caption*{\uapo}
	\end{subfigure}%
	\hspace{1.0em}%
	\begin{subfigure}{.13\linewidth}
		\centering
		\includegraphics[width = \linewidth]{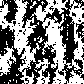}
		\caption*{\osfwu}
	\end{subfigure}
	\caption{Comparison of the perturbation size added into the same clean (RGB) observation in different attacks against the DQN agent playing Pong and $\epsilon=0.01$. In perturbations, black pixels: $-0.01$, white pixels: $+0.01$, gray pixels: $0.0$.}\label{fig:distortionPong}
\end{figure*}

Figure~\ref{fig:distortionPong} shows that $\vect{r}$ obtained via \uaps and \uapo are smaller than other adversarial perturbations for the same $\epsilon$, since \uaps and \uapo try to find a minimal $\vect{r}$ that sends all $\vect{x} \in \mathcal{D}_{train}$ outside the decision boundary~\cite{moosavi2017universal}.
We conclude that \uaps and \uapo are likely to be less detectable based on the amount of the perturbation (e.g., via visual observation). Appendix~\ref{sec:adv_perturb_appendix} also presents additional results for other agents and other games, as well as the perturbed version of the observations.

As summarized in Table~\ref{tab:attack_summary}, FGSM computes a new $\vect{r}$ for each $\vect{s}$ after observing the complete $\vect{s}$. Therefore, it requires rewriting \victim's memory to change all previously stored observations $o_j$ of the current \s, in which \victim is attacked. Unlike FGSM, \uaps and \uapo add $r_j$ to the incoming $o_j$, and \victim stores adversarially perturbed $o_j$ into its memory. The online cost of \osfw is too high, and it cannot be mounted without interfering with the environment. 
\uaps, \uapo and \osfwu are real-time attacks that do not require stopping the agent or the environment while adding $\vect{r}$. \uaps and \osfwu generate $\vect{r}$ that is independent of $\vect{s}$, but the $r_j$ for each observation $o_{j} \in \vect{s}$ is different. On the other hand, \uapo adds the same $r_{j}$ to all observations in any $\vect{s}$, which makes the perturbation generation independent of the size of $\vect{s}$.
\uapo leads to an efficient and effective attack when $\epsilon \geq 0.006$, and does not have temporal and observational dependency. \uaps is the optimal attack considering both effectiveness and efficiency.

\begin{table}[t]
	\begin{center}
		\resizebox{0.92\columnwidth}{!}{
			\begin{tabular}{lccc}
				\hline
				\bf Attack & \bf Online \bf & \bf State  & \bf Observation \\ 
				\bf Method & \bf cost & \bf dependency & \bf dependency \\ \hline
				FGSM~\cite{huang2017adversarial} &  Low & Dependent & Dependent\\
				\osfw~\cite{xiao2019characterizing} & High & Independent & Dependent\\
				\underline{\textcolor{blue}{\uaps}} & Low & Independent & Dependent \\
				\underline{\textcolor{blue}{\uapo}} & Low & Independent & Independent\\
				\underline{\textcolor{blue}{\osfwu}} & Low & Independent & Dependent\\
				\hline
			\end{tabular}}
		\end{center}
		\caption{Summary of five attacks based on the characteristics that makes an attack plausible in a real deployment scenario. New attacks proposed in this paper are highlighted in blue and underlined.}\label{tab:attack_summary}
	\end{table}

In simple environments like Atari 2600 games, sequential states are not i.i.d. Moreover, Atari 2600 games are \emph{controlled} environments, where future states are predictable and episodes do not deviate much from one another for the same task. \osfw and \osfwu leverage this non-i.i.d property. However, their effectiveness might decrease in uncontrolled environments and the physical world due to the uncertainty of the future states.
In contrast, \uaps and \uapo are independent of the correlation between sequential states. To confirm our conjecture, we implemented \osfwu against VGG-16 image classifers~\cite{zhang2015accelerating} pre-trained on ImageNet~\cite{russakovsky2015imagenet}, where ImageNet can be viewed as a non i.i.d., uncontrolled environment. We measured that \osfwu achieves a fooling rate of up to $30\%$ on the ImageNet validation set with $\epsilon=10$, while \uaps has a fooling rate of $78\%$~\cite{moosavi2017universal}, where the $l_{\infty}$ norm of an image in the validation set is around 250. Therefore, we conclude that \uaps can mislead DRL policies more than \osfwu in complex, uncontrolled environments. 

\subsection{Attack Performance in Continuous Control}\label{ssec:performancecontinuous}

\begin{figure}[t]
	\Description[<short description>]{<long description>}
    \centering
    \includegraphics[width=1.0\columnwidth]{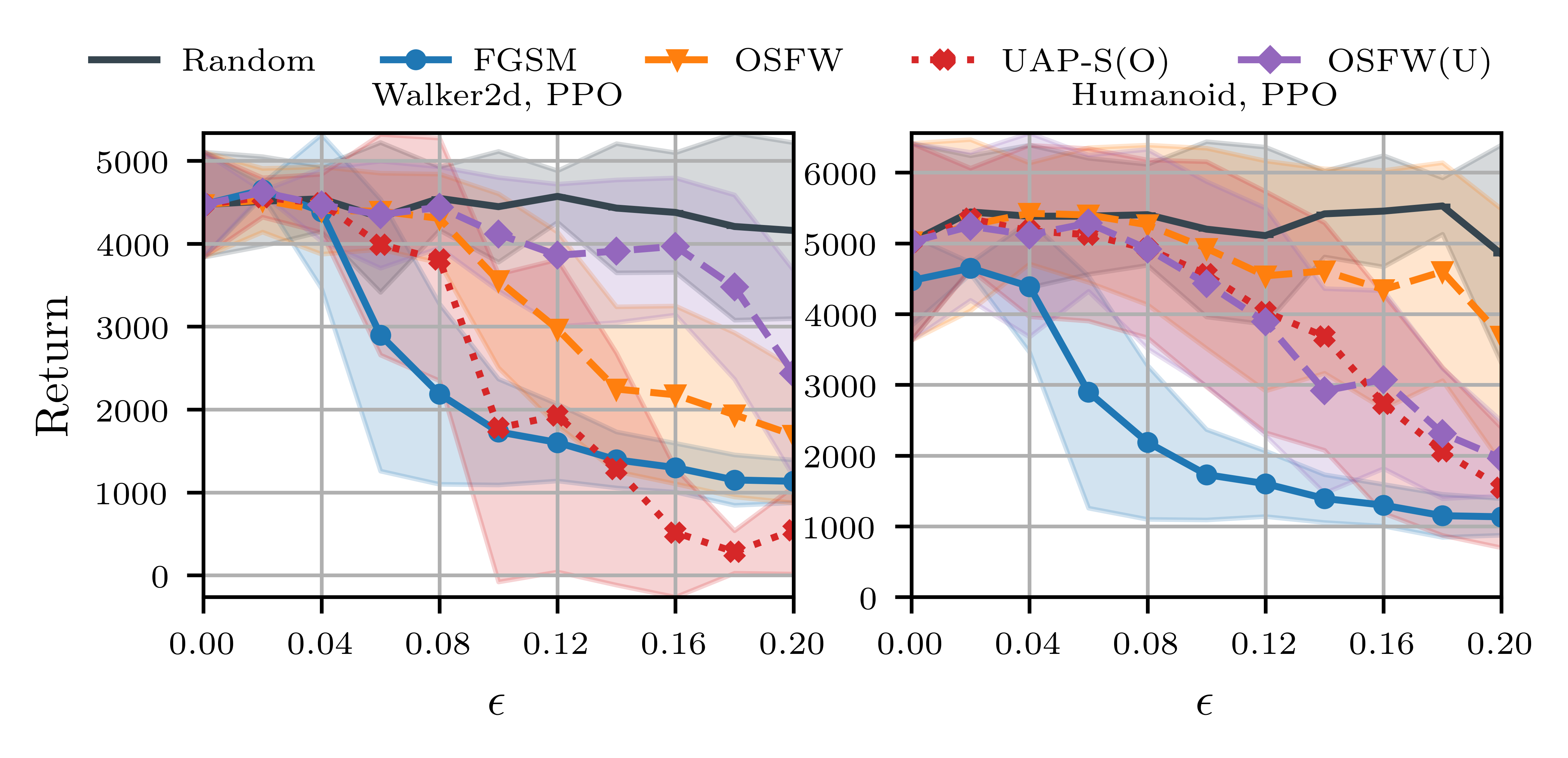}
    \caption{Comparison of attacks against PPO agents trained for Humanoid and Walker-2d tasks. The graph shows how the return (averaged over 50 games) changes with different $\epsilon$ values for five different attack strategies.} \label{fig:attackcomparison_continuouscontrol}
\end{figure}

\begin{table}[t]
    \centering
    \resizebox{1.0\columnwidth}{!}{
    	\begin{tabular}{llcc}
    		\hline
    		\bf \multirow{2}{*}{Experiment} & \bf Attack & \bf Offline cost$\pm$ std & \bf Online cost $\pm$ std  \\ 
    		& \bf method & \bf (seconds) & \bf (seconds) \\ \hline
    		\multirow{5}{0.4\linewidth}{
    			Walker2d, PPO, $T_{max} = 0.0079\pm10^{-5}$ seconds
    		} & FGSM &  - &   $31\times10^{-5}\pm10^{-5}$ \\
    		& \osfw & - & \textcolor{red}{ $0.02 \pm 0.001$} \\
    		& \uaps(O) & $8.75 \pm 0.024 $ & $2.9\times10^{-5}\pm10^{-6}$ \\
    		& \osfwu & $0.02 \pm 0.001$ &  $2.9\times10^{-5}\pm10^{-6}$  \\ \hline
    		\multirow{5}{0.4\linewidth}{
    			Humanoid PPO, $T_{max} = 0.0079\pm10^{-6}$ seconds}
    		& FGSM & - & $35\times10^{-5}\pm10^{-5}$\\
    		& \osfw & - & \textcolor{red}{$0.02 \pm 0.001$} \\
    		& \uaps(O) & $35.86 \pm 0.466 $  & $2.4\times10^{-5}\pm10^{-6}$\\
    		& \osfwu & $0.02 \pm 0.001$ &  $2.4\times10^{-5}\pm10^{-6}$\\ \hline
    	\end{tabular}}
    	    \caption{Offline and online cost of attacks
    	    	where victim agents are PPO trained for Walker2d and Humanoid at $\epsilon=0.02$. Attacks that cannot be implemented in real-time are highlighted in red.}
    	    \label{tab:testtiming_comparison_continuous}
\end{table}

In continuous control, we used PPO agents in~\cite{zhang2020robust} pre-trained for two different MuJoCo tasks (Walker2d and Humanoid) as \victim\footnote{Agents are downloaded from \url{https://github.com/huanzhang12/SA_PPO} and frame rates are set to default values as in \url{https://github.com/openai/gym}}. We used the original experimental setup to compare our attacks with baseline attacks and random noise addition. In our experiments, PPO agents show performance similar to those reported in the original paper~\cite{zhang2020robust}.
We implemented \uaps by copying the parameters of $V_{\mathit{v}}$ into $V_{\mathit{adv}}$, and set $\delta_{max}=95\%$. FGSM, \osfw and \osfwu also use $V_{\mathit{v}}$ to minimize the value in a perturbed state.
Additionally, in both tasks, $\vect{s}$ contains only one observation, which reduces \uaps to \uapo. 

Figure~\ref{fig:attackcomparison_continuouscontrol} shows the attack effectiveness when $0.0 \leq \epsilon \leq 0.2$. FGSM is the most effective attack in Humanoid, while \uaps decreases the return more than FGSM in Walker2d when $\epsilon \geq 0.12 $. Overall, all attacks behave similarly in both discrete and continuous action spaces, and our conclusions regarding the effectiveness of universal adversarial perturbations generalize to continuous control tasks, where $Q_{\mathit{v}}$ is not available. However, in these tasks, \adversary only decreases the critic's evaluation of $\vect{s}$ when taking an action $a$. Even if \adversary decreases the value of $V_{\mathit{adv}}(\vect{s})$, it does not necessarily lead to $\pi_{\mathit{v}}$ choosing a sub-optimal action over the optimal one. Therefore, attacks using $V$ require $\epsilon \geq 0.2$ to fool $\pi_{\mathit{v}}$ effectively. For a more efficient attack, \adversary can copy $\pi_{\mathit{v}}$ into a proxy agent as $\pi_{\mathit{adv}}$, and try to maximize the total variation distance~\cite{zhang2020robust} in $\pi_{\mathit{adv}}$ for states perturbed by universal perturbations. \uaps, \uapo and \osfwu need to be modified to compute the total variation distance, and we leave this as a future work.

Table~\ref{tab:testtiming_comparison_continuous} presents the online and offline computational cost for the perturbation generation. The results are comparable with Table~\ref{tab:testtiming_comparison}, and confirm that FGSM, \uaps, \uapo and \osfwu can be mounted in real-time, although FGSM needs write-access to the agent's memory. \osfw cannot inject the generated noise immediately into subsequent states, since it has a higher online cost than the maximum upper bound. 


\begin{table*}[t]
	\begin{center}
		\begin{subtable}[t]{0.90\textwidth}
			\centering
			\resizebox{\textwidth}{!}{
				\begin{tabular}{lccccccc}
					\hline
					& & \multicolumn{6}{c}{\bf Average return $\pm$ std in the presence of adversarial perturbation attacks} \\
					\bf {epsilon} & \bf Defense & \bf No attack & \bf FGSM & \bf \osfw &  \bf \uaps & \bf \uapo &  \bf \osfwu \\ \hline
					\multirow{4}{*}{$0.01$}  & No defense & $21.0\pm0.0$ &  $\bm{-21.0\pm0.0}$ & $-20.0\pm3.0$ & $\bm{-21.0\pm0.0}$ & $-19.8\pm0.4$ & $\bm{-21.0\pm0.0}$ \\
					& \vf~\cite{lin2017detecting}  & \colorbox{CGreen}{$21.0\pm0.0$} & \colorbox{CGreen}{$21.0\pm0.0$} & $-19.7\pm0.5$ & \colorbox{CGreen}{$0.7\pm1.7$} & $0.4\pm2.7$ &   $\bm{-21.0\pm0.0}$ \\
					& SA-MDP~\cite{zhang2020robust}  & \colorbox{CGreen}{$21.0\pm0.0$} & \colorbox{CGreen}{$21.0\pm0.0$} & \colorbox{CGreen}{$21.0\pm0.0$} & \colorbox{CGreen}{$21.0\pm0.0$} & \colorbox{CGreen}{$21.0\pm0.0$} & \colorbox{CGreen}{$21.0\pm0.0$} \\
					\hline
					\multirow{4}{*}{$0.02$}  & No defense & $21.0\pm0.0$ & $-19.9\pm1.3$ & $\bm{-21.0\pm0.0}$ & $-20.8\pm0.6$ & 
					$-20.0\pm0.0$ & $\bm{-21.0\pm0.0}$ \\
					& \vf~\cite{lin2017detecting}  & \colorbox{CGreen}{$21.0\pm0.0$} & \colorbox{CGreen}{$21\pm0.0$} & \colorbox{CBlue}{$-19.7\pm0.6$} & \colorbox{CBlue}{$9.4\pm0.8$} & \colorbox{CBlue}{$5.3\pm3.9$} & \colorbox{CBlue}{$-20.5\pm0.5$} \\
					& SA-MDP~\cite{zhang2020robust}  & \colorbox{CGreen}{$21.0\pm0.0$} & $-14.6\pm8.8$ & $-20.5\pm0.5$ &  $-20.6\pm0.5$ & $-20.6\pm0.5$ &$\bm{-21.0\pm0.0}$ \\
					\hline
					\multirow{4}{*}{$0.05$}  & 
					No defense & $ 21.0\pm0.0$ & $-20.5\pm 0.7 $ & $\bm{-21.0\pm 0.0}$ & $-20.6\pm0.8$ & $-20.0\pm0.0$ & $\bm{-21.0\pm0.0}$  \\
					& \vf~\cite{lin2017detecting}  & \colorbox{CGreen}{$21.0\pm 0.0$}& \colorbox{CGreen}{$21.0\pm0.0$}  & \colorbox{CBlue}{$-20.0\pm 0.0$} & \colorbox{CBlue}{$7.6\pm 4.7$} & \colorbox{CBlue}{$-14.1\pm 1.1$} & $\bm{-21.0\pm 0.0}$ \\
					& SA-MDP~\cite{zhang2020robust}  & \colorbox{CGreen}{$21.0\pm 0.0$} & $\bm{-21.0\pm 0.0}$ &  $\bm{-21.0\pm0.0}$ & $-20.6\pm 0.5$ & $-20.6\pm 0.5$ & $\bm{-21.0\pm0.0}$\\
					\hline
				\end{tabular}}
				\caption{DQN agent playing Pong}
			\end{subtable}
			\hfill
			\begin{subtable}[t]{0.90\textwidth}
				\centering
				\resizebox{\textwidth}{!}{
					\begin{tabular}{lccccccc}
						\hline
						& & \multicolumn{6}{c}{\bf Average return $\pm$ std in the presence of adversarial perturbation attacks} \\
						\bf {epsilon} & \bf Defense & \bf No attack & \bf FGSM & \bf \osfw &  \bf \uaps & \bf \uapo &  \bf \osfwu \\ \hline
						\multirow{4}{*}{$0.01$}  & No defense &  $34.0\pm0.0$ &   $\bm{0.0\pm0.0}$& $2.0\pm1.1$ &  $2.1\pm0.8$& $4.0\pm0.6$ & $0.5\pm0.5$\\
						& \vf~\cite{lin2017detecting}  & \colorbox{CGreen}{$32.0\pm1.5$} & \colorbox{CGreen}{$32.6\pm1.7$} & $24.1\pm1.0$ & $22.9\pm0.9$ & $25.8\pm1.1$ & $\bm{20.9\pm1.2}$\\
						& SA-MDP~\cite{zhang2020robust}  & $30.0\pm0.0$ & $30.0\pm0.0$ & \colorbox{CGreen}{$30.0\pm0.0$} & \colorbox{CGreen}{$30.0\pm0.0$} & \colorbox{CGreen}{$30.0\pm0.0$} & \colorbox{CGreen}{$30.0\pm0.0$} \\
						\hline
						\multirow{4}{*}{$0.02$}  & No defense & $34.0\pm0.0$  & $\bm{0.0\pm0.0}$& $1.0\pm0.0$ & $0.1\pm0.3$& $0.8\pm0.6$& 
						$\bm{0.0\pm0.0}$\\
						& \vf~\cite{lin2017detecting}  & \colorbox{CGreen}{$32.0\pm1.5$} & \colorbox{CGreen}{$32.6\pm1.7$} &  $\bm{1.1\pm0.3}$ & $24.0\pm2.0$ & $25.6\pm1.0$  & $4.4\pm1.1$ \\
						& SA-MDP~\cite{zhang2020robust}  & $30.0\pm0.0$ &  $29.8\pm0.6$  & \colorbox{CGreen}{$29.9\pm0.3$}  & \colorbox{CGreen}{$\bm{29.4\pm1.2}$} & \colorbox{CGreen}{$\bm{29.4\pm1.2}$} &  \colorbox{CGreen}{$30.0\pm0.0$}\\
						\hline
						\multirow{4}{*}{$0.05$}  & No defense 
						& $34.0\pm0.0$  & $\bm{0.0 \pm 0.0}$ & $1.2\pm 0.0$ & $2.2 \pm 1.7$ & $2.2 \pm 1.4$ & $\bm{0.0\pm0.0}$\\
						& \vf~\cite{lin2017detecting}  & \colorbox{CGreen}{$32.0\pm 1.4$} & \colorbox{CGreen}{$32.6\pm 1.6$}  & $1.0\pm 0.0$ & \colorbox{CBlue}{$29.0 \pm 1.1 $}& \colorbox{CBlue}{$23.9\pm 0.3$} &  $\bm{0.0 \pm 0.0}$\\
						& SA-MDP~\cite{zhang2020robust}  & $30.0 \pm 0.0$ & $21.1\pm 1.3$ & \colorbox{CBlue}{$\bm{20.9\pm 0.8}$} & $21.1\pm 1.7$ & $21.1\pm 1.7$  & \colorbox{CBlue}{$21.1 \pm 1.7$}\\
						\hline
					\end{tabular}}
					\caption{DQN agent playing Freeway}
				\end{subtable}
			\end{center}
			\caption{Average returns (10 episodes) for the DQN agent playing Pong in the presence of different adversarial perturbations, and agents are equipped with different defenses. In each row, the best attack (lowest return) is in bold. In each column, for a given $\epsilon$ value, the most robust defense (highest return) for that particular attack is shaded green if the defense can fully recover the victim's return, and blue if the victim's return is not fully recovered.}\label{tab:defenses}
		\end{table*}

\section{Detection and Mitigation of Adversarial Perturbations}\label{sec:detectingandmitigating}

A defender's goal is the opposite of the adversary \adversary's goal.
Section~\ref{sec:advModel} outlines three criteria for a successful attack. A good defense mechanism should thwart one or more of these criteria.
We begin by surveying previously proposed defenses (Section~\ref{ssec:defensemethods}), followed by an evaluation of the effectiveness of two of the most relevant defenses (Section~\ref{subsec:defense-effectiveness}) that aim to address the first (lowering the victim \victim's return) and third attack criterion (evasion of detection).
We then present a new approach to detect the presence of adversarial perturbations in state observations (Section~\ref{subsec:ad3}), aimed at thwarting third attack criterion on evading defenses.

\subsection{Defenses against Adversarial Examples}\label{ssec:defensemethods}

\subsubsection{Defenses in image classification} 
Many prior studies proposed different methods to differentiate between normal images and adversarial examples. Meng and Chen~\cite{meng2017magnet} or Rouhani \textit{et al.}~\cite{rouhani2018deepfense} model the underlying data manifolds of the normal images to differentiate clean images from adversarial examples. Xu \textit{et al.}~\cite{xu2017feature} propose feature squeezing to detect adversarial examples. However, these detection methods are found to be inefficient~\cite{carlini2017adversarial} and not robust against adaptive adversaries that can specifically tailor adversarial perturbation generation methods for a given defense~\cite{tramer2020adaptive}.

\subsubsection{Defenses in DRL} 

Previous work on adversarial 
training~\cite{behzadan2017whatever,kos2017delving} presents promising results as a defense. However, Zhang et al.~\cite{zhang2020robust} show that adversarial training leads to unstable training, performance degradation, and is not robust against strong attacks. Moreover, Moosavi et al.~\cite{moosavi2017universal} prove that despite a slight decrease in the fooling rate $\delta$ in the test set, \adversary can easily compute another universal perturbation against retrained agents. 
To overcome the challenges of adversarial training, Zhang et al.~\cite{zhang2020robust} propose state-adversarial Markov decision process (SA-MDP), which aims to find an optimal $\pi$ under the strongest \adversary using policy regularization. This regularization technique helps DRL agents \emph{maintain} their performance even against adversarially perturbed inputs. Similarly, Oikarinen et al.~\cite{oikarinen2021robust} use adversarial loss functions during training to improve the robustness of agents.

Visual Foresight (\vf)~\cite{lin2017detecting} is another defense that \emph{recovers} the performance of an agent in the presence of \adversary. \vf predicts the current observation $\hat{o}_{t}$ at time $t$ using $k$ previous observations $o_{t-k}:o_{t-1}$ and corresponding actions $a_{t-k}:a_{t-1}$. It also predicts the possible action $\hat{a}_{t}$ for the partially predicted $\hat{\vect{s}}$ using $Q(\hat{\vect{s}},a)$.
The difference between $Q(\hat{\vect{s}},a)$ and $Q(\hat{\vect{s}},\hat{a})$ determines whether \s is perturbed or not. In the case of detection, $\hat{a}_{t}$ is selected to recover the performance.

\subsection{Effectiveness of Existing Defenses}
\label{subsec:defense-effectiveness}

To investigate the limitations of previously proposed defenses for DRL, we implemented two defense methods that aim to retain the average return of \victim when it is under attack: \vf~\cite{lin2017detecting} and SA-MDP~\cite{zhang2020robust}, both of which seek to prevent the first attack objective that limits \victim to a low return. \vf also prevents the third attack criterion and detects adversarial perturbations.
Since we want to evaluate the effectiveness of \vf and SA-MDP, we focus on the DQN agents for Pong and Freeway as these are the ones that are common between our experiments (Section~\ref{sec:Evaluation}) and these defenses (\cite{lin2017detecting,zhang2020robust}).

We implemented \vf from scratch for our DQN models following the original experimental setup in~\cite{lin2017detecting} by setting $k=3$ to predict every $4^{th}$ observation. We also set the pre-defined threshold value to $0.01$, which is used to detect adversarial perturbations to achieve the highest detection rate and performance recovery. We downloaded state adversarial DQN agents, which are trained using SA-MDP,
from their reference implementation\footnote{\url{https://github.com/chenhongge/SA_DQN} Downloaded SA-MDP agents use $\epsilon=1/255$ in training as in the original work~\cite{zhang2020robust}. Using higher $\epsilon$ values in training led to poor performance.}. SA-MDP agents only use one observation per $\vect{s}$; therefore, UAP-S would reduce to UAP-O for SA-MDP in this setup.
Table~\ref{tab:defenses} shows the average return for each agent under a different attack, and while equipped with different defenses.
In Section~\ref{sec:Evaluation}, we established that when the perturbation bound $\epsilon$ is $0.01$, all attacks are devastatingly effective. In the interest of evaluating the robustness of the defense, we also consider two higher $\epsilon$ values, $0.02$ and $0.05$.

\vf is an effective defense against FGSM and \uaps as it can recover \victim's average return. However, it is not very effective against \osfw, \osfwu, and \uapo. 
SA-MDP is better than \vf against \osfw and \osfwu when $\epsilon = 0.01$, but it fails to defend any attack in Pong when $\epsilon \geq 0.02$. Notably, \vf's detection performance depends on the accuracy of the action-conditioned frame prediction module used in its algorithm~\cite{lin2017detecting}, and the pre-defined threshold value used for detecting adversarial perturbations. SA-MDP is able to partially retain the performance of \victim when $\epsilon = 0.01$ and $0.02$ in the case of Freeway. Notably, \victim's average return when it is under attack is still impacted, and SA-MDP cannot retain the same average return as when there is no attack.
Furthermore, SA-MDP's effectiveness decreases drastically when the perturbation bound $\epsilon$ is increased to $0.05$.

\subsection{Action Distribution Divergence Detector(\detector)}
\label{subsec:ad3}

\subsubsection{Methodology} 
In a typical DRL episode, sequential actions exhibit some degree of \emph{temporal coherence}: the likelihood of the agent selecting a specific current action given a specific last action is consistent across different episodes. We also observed that the temporal coherence is disrupted when the episode is subjected to an attack. 
We leverage this knowledge 
to propose a detection method, Action Distribution Divergence Detector (\detector), which calculates the statistical distance between the \emph{conditional action probability distribution} (CAPD) of the current episode to the learned CAPD in order to detect whether the agent is under attack.
Unlike prior work on detecting adversarial examples in the image domain~\cite{meng2017magnet,rouhani2018deepfense,xu2017feature}, \detector does not analyze the input image or tries to detect adversarial examples. Instead, it observes the distribution of the actions triggered by the inputs and detects unusual action sequences.
To train \detector, \victim first runs $k_1$ episodes in a controlled environment before deployment. \detector saves all actions taken during that time and approximates the conditional probability of the next action given the current one using the bigram model\footnote{We tested different ngrams and selected the bigram as the best option.}.
We call the conditional probability of actions approximated by $k_1$ episodes the \emph{learned} CAPD.
Second, to differentiate between the CAPD of a normal game versus a game that is under attack with high confidence,
$v$ runs another $k_2$ episodes in a safe environment. \detector decides a threshold value $th$, where the statistical distance between the CAPD of the normal game and the learned CAPD falls mostly below this threshold.
We use Kullback-Leibler (KL) divergence~\cite{kl-div} as the statistical distance measure.
The KL divergence between the learned CAPD and the CAPD of the current episode is calculated at each time-step starting after the first $t_1$ steps.
We skip the first $t_1$ steps because the CAPD of the current episode is initially unstable and the KL divergence is naturally high at the beginning of every episode.
We set the threshold $th$ as the $p^{th}$ percentile of all KL-divergence values calculated for $k_2$ episodes. 
During deployment, \detector continuously updates the CAPD of the current episode, and after $t_1$ steps, it calculates the KL-divergence between the CAPD of the current episode and the learned CAPD. If the KL-divergence exceeds the threshold $th$ by $r$\% or more during a time window $t_2$, then \detector raises an alarm that the agent is under attack.

\subsubsection{Evaluation} We evaluate the precision and recall of \detector in three tasks with discrete action spaces (Pong, Freeway and Breakout) against all proposed attacks.
\detector can detect all five attacks in Pong with perfect precision and recall scores in all configurations. In Freeway, \detector has perfect precision and recall scores for 12 out of 15 different setups (FGSM against the DQN agent, \osfw against the PPO agent, and \osfwu against the PPO agent). In Freeway, we found that attacks lead to a lower action change rate (as low as $20\% - 30\%$ in some episodes) than other tasks, thus negatively affecting the precision and/or recall of \detector. \detector is also less effective in Breakout, as this task often terminates too quickly when it is under attack, and \detector cannot store enough actions for CAPD to converge in such a short time. The optimal parameters used for training \detector, a more detailed discussion of the performance of \detector, and the full result of the evaluation can be found in Appendix~\ref{sec:additionalexp_appendix}.

\subsubsection{Limitations} 
\detector is designed for attacks where \adversary injects adversarial perturbations consistently throughout the episode. \adversary with the knowledge of the detection strategy could apply their adversarial perturbation at a lower frequency to avoid detection. However, lowering the attack frequency also decreases the attack effectiveness. Another way for \adversary to evade \detector is to perform targeted attacks to lure \victim into a specific state, where the adversarial perturbation is applied to a limited number of states in an episode. This type of attack is outside the scope of our paper, and defense strategies against it will be explored in future work.

\subsubsection{Adversary vs. defender strategy with negative returns}

\begin{table}[t]
	\centering
	\resizebox{1.0\columnwidth}{!}{
		\begin{tabular}{lccccccc}
			\hline
			& & \multicolumn{6}{c}{\bf Losing Rate} \\
			\bm$\epsilon$ & \bf Method & \bf No attack & \bf FGSM & \bf \osfw & \bf \uaps & \bf \uapo & \bf \osfwu\\ \hline
			\multirow{3}{*}{$0.01$}  & No defense & 0.0 & \boldmath $1.0$ & \boldmath $1.0$ & \boldmath $1.0$ & \boldmath $1.0$ & \boldmath $1.0$ \\ 
			& \vf~\cite{lin2017detecting} & 0.0 & 0.0 & \colorbox{CRed}{\boldmath $1.0$} & 0.0 & \colorbox{CRed}{0.2} &  \colorbox{CRed}{\boldmath $1.0$}  \\
			& SA-MDP~\cite{zhang2020robust} & 0.0 & 0.0 & 0.0 & 0.0 & 0.0 & 0.0 \\ 
			& \detector & 0.0 & 0.0 & 0.0 & 0.0 & 0.0 & 0.0 \\ \hline
			\multirow{3}{*}{$0.02$} & No defense & 0.0 & \boldmath $1.0$ & \boldmath $1.0$ & \boldmath $1.0$ & \boldmath $1.0$ & \boldmath $1.0$ \\ 
			& \vf~\cite{lin2017detecting} & 0.0 & 0.0 & \colorbox{CRed}{\boldmath $1.0$} & 0.0 & 0.3 & \colorbox{CRed}{\boldmath $1.0$}  \\
			& SA-MDP~\cite{zhang2020robust} & 0.0 & \colorbox{CRed}{ $0.9$} & \colorbox{CRed}{\boldmath $1.0$} & \colorbox{CRed}{\boldmath $1.0$}  & \colorbox{CRed}{\boldmath $1.0$}  & \colorbox{CRed}{\boldmath $1.0$}  \\ 
			& \detector & 0.0 & 0.0 & 0.0 & 0.0 & 0.0 & 0.0 \\ \hline
		\end{tabular}}
		\caption{Losing rate (10 episodes) of DQN agents playing Pong with or without additional defense. Losing rate is calculated by counting the number of games where the computer gains 21 points first in an episode. If \detector raises an alarm before an episode ends, then \victim does not lose the game. In each row, the best attack with the highest losing rate is in bold, and given an $\epsilon$ value, the defense with the highest losing rate for that particular attack is shaded red.}\label{tab:losegames}
	\end{table}
	
In any DRL task where there is a clear negative result for an episode (e.g., losing a game) or the possibility of negative return, a reasonable choice for \victim is to suspend an episode when \adversary's presence is detected.
For example, in Pong, a negative result would be when the computer (as the opponent) reaches the score of 21 before \victim does. Suspending an episode prevents \victim from losing the game. Defense mechanisms such as \vf and SA-MDP are useful in retaining or recovering \victim's return; however, they may not always prevent \victim from falling into a negative result, e.g., losing the game in Pong. Combining a recovery/retention mechanism with suspension on attack detection can reduce the number of losses for \victim. 

To illustrate the effectiveness of combining  \detector with a retention/recovery mechanism, we designed an experiment using a DQN agent playing Pong to compare \emph{losing rate} of \victim when it is under attack. We used Pong for this experiment, as Pong has a clear negative result.
In Pong, an episode ends with loss when (a) the computer reaches 21 points before \victim, or (b) \detector did not raise an alarm. The result of this experiment can be found in Table~\ref{tab:losegames}. As shown in this table, \vf is not effective in reducing the losing rate of \victim for \osfw and \osfwu for all $\epsilon$. SA-MDP is effective in avoiding losses when $\epsilon = 0.01$; however, it fails against all universal perturbations when $\epsilon = 0.02$. In contrast, \detector can detect the presence of adversarial perturbations in all games. Although retention/recovery and detection are two orthogonal aspects of defense, our results above suggest that they can be combined in tasks with negative returns or results in order to more effectively thwart \adversary from achieving its first goal. In the scenario where the defense mechanism fails to recover enough reward for the \victim to avoid a negative end state, the detection module can be used as a forensic method to understand where and why the system fails and to forfeit negative results when \adversary is present.

\section{Conclusion}
\label{sec:Conclusion}
We showed that white-box universal perturbation attacks are effective in fooling DRL policies in real-time. Our evaluation of the three different attacks (\uaps, \uapo, and \osfwu) demonstrates that universal perturbations are effective in tasks with discrete action spaces. Universal perturbation attacks are also able to generalize to continuous control tasks with the same efficiency. We confirmed that the effectiveness of prior defenses depends on the perturbation bound, and fail to completely recover the agent performance when they are confronted with universal perturbations of larger bounds.
We proposed a detection mechanism, \detector, that detects all five attacks evaluated in the paper. \detector can be combined with other defense techniques to protect agents in tasks with negative returns or results to stop the adversary from achieving its goal.
We plan to extend our attacks to the black-box case by first mounting a model extraction attack and then applying our current techniques to find transferable universal perturbations.

\subsubsection*{Acknowledgements} This research was funded in part by the EU H2020 project SPATIAL (Grant No. 101021808) and Intel Private-AI Consortium.


\bibliographystyle{acm}
\bibliography{bibliography}

\appendix

\section{Proof of Equation~\ref{eqn:uapo}}\label{sec:proof}
In this section, we provide a formal proof of Equation~\ref{eqn:uapo} that was used to find the perturbation in \uapo.

In \uapo, we defined an additional constraint for the perturbation as $\tilde{r}_{j}= \tilde{r}_{k},\; \forall j,k \in \{t-N+1, \cdots, t \}$, so that the same $\tilde{r}_{j}$ is applied into all observations $o_j$. As pointed out in Section~\ref{subsec:UAPAttack}, at $i$-th iteration, the DeepFool algorithm computes the $i$-th perturbation $\vect{r}_i$ for $i$-th state $\vect{s}_i$ as
\begin{equation}
    \Delta \vect{r}_i \gets 
    \frac{| Q^{'}(\vect{s}^{*}_i, a_{\hat{l}}) |}{ {\lVert {\vect{w}^{'}_{\hat{l}}} \rVert}^{2}_{2} }\vect{w}^{'}_{\hat{l}}.
\end{equation}

We can keep the term $| Q^{'}(\vect{s}^{*}_i, a_{\hat{l}}) |/{ {\lVert {\vect{w}^{'}_{\hat{l}}} \rVert}^{2}_{2}}$ as constant, since it gives the magnitude of the perturbation update $\Delta \vect{r}_i$, and we can maintain the same magnitude of $\Delta \vect{r}_i$ in $\Delta \tilde{\vect{r}}_i$. Then, in \uapo, DeepFool finds the closest $\Delta \tilde{\vect{r}}_i$ to $\Delta \vect{r}_i$ by minimizing the direction of the perturbation as ${\lVert {\vect{w}^{'}_{\hat{l}} - \Delta \tilde{\vect{r}}_i} \rVert}^{2}_{2}$. 

We can find the optimal $\Delta \tilde{\vect{r}}_i$ by setting $\partial{\lVert {\vect{w}^{'}_{\hat{l}} - \Delta \tilde{\vect{r}}_{i}} \rVert}_{2}/{\partial \Delta \tilde{r}_{i_j}}$ to zero for all $j \in \{t-N+1, \cdots, t\}$, where $N$ denotes the total number of observations inside state $\vect{s}$. We also set all $\Delta \tilde{r}_{i_j}$ into the same variable $\tilde{r}_{i_t}$. Therefore, we simply obtain $\Delta \tilde{r}_{i_t}$ by finding the point where the partial derivative is equal to zero as in Equation~\ref{eqn:squaredderivate}. 
\begin{equation}\label{eqn:squaredderivate}
\begin{split}
   \frac{\partial{\lVert {\vect{w}^{'}_{\hat{l}} - \Delta \tilde{r}_{i_t}} \rVert}_{2}}{{\partial \Delta \tilde{r}_{i_t}}} &= \frac{\partial ((\vect{w}^{'}_{\hat{l}_{t}} - \Delta \tilde{r}_{i_t})^2 + \cdots +  (\vect{w}^{'}_{\hat{l}_{t-N}} - \Delta \tilde{r}_{i_t})^2)^{1/2}}{{\partial \Delta \tilde{r}_{i_t}}} \\
    0 & = \frac{\partial f(\Delta \tilde{r}_{i_t})^{1/2}}{{\partial \Delta \tilde{r}_{i_t}}} \\
    0 & = 0.5f(\Delta \tilde{r}_{i_t})^{-3/2}
    \Big(2(N+1)\Delta \tilde{r}_{i_t} - 2\; \smashoperator{\sum^{t}_{k=t-N+1}}\vect{w}^{'}_{\hat{l}_k}
    \Big) \\
\end{split}
\end{equation}

The squared difference function $f(\Delta \tilde{r}_{i_t})^{-3/2} \geq 0$, and only equal to zero when $\Delta \tilde{r}_{i_t} = 0$, which does not give any useful perturbation. 
Therefore, Equation~\ref{eqn:squaredderivate} is solved when $\vect{w}^{'}_{\hat{l}_k} = \Delta \tilde{r}_{i_t},\; \forall k \in \{t-N+1, \cdots, t\}$. Therefore, by putting back the constant, DeepFool computes the perturbation at $i^{th}$ iteration as 
\begin{equation}
    \Delta \tilde{r}_{i_j}\gets  \frac{| Q^{'}(\vect{s}^{*}_i, a_{\hat{l}}) |}{ N{\lVert {\vect{w}^{'}_{\hat{l}}} \rVert}^{2}_{2} }\quad \smashoperator{\sum_{k=(t-N+1)}^{t}}\;\vect{w}^{'}_{\hat{l}_{k}}.
\end{equation} 

\section{Adversarial Perturbations}
\label{sec:adv_perturb_appendix}

\begin{figure*}[t]
	\Description[<short description>]{<long description>}
  \centering
  \begin{subfigure}{.13\linewidth}
    \centering
    \includegraphics[width = \linewidth]{figures/distortions/Pong_clean_rgb.png}
  \end{subfigure}%
  \hspace{0.5em}%
  \begin{subfigure}{.13\linewidth}
    \centering
    \includegraphics[width = \linewidth]{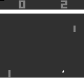}
  \end{subfigure}%
  \hspace{0.5em}%
  \begin{subfigure}{.13\linewidth}
    \centering
    \includegraphics[width = \linewidth]{figures/distortions/Pong_fgsm_perturbation.png}
  \end{subfigure}%
  \hspace{0.5em}%
   \begin{subfigure}{.13\linewidth}
    \centering
    \includegraphics[width = \linewidth]{figures/distortions/Pong_obs_fgsm_wb_ingame_perturbation.png}
  \end{subfigure}%
  \hspace{0.5em}%
  \begin{subfigure}{.13\linewidth}
    \centering
    \includegraphics[width = \linewidth]{figures/distortions/Pong_uap_s_perturbation.png}
  \end{subfigure}
  \hspace{0.5em}%
  \begin{subfigure}{.13\linewidth}
    \centering
    \includegraphics[width = \linewidth]{figures/distortions/Pong_uap_f_perturbation.png}
  \end{subfigure}%
  \hspace{0.5em}%
  \begin{subfigure}{.13\linewidth}
    \centering
    \includegraphics[width = \linewidth]{figures/distortions/Pong_obs_fgsm_wb_perturbation.png}
  \end{subfigure}

  \begin{subfigure}{.13\linewidth}
    \centering
    \includegraphics[width = \linewidth]{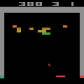}
  \end{subfigure}%
  \hspace{0.5em}%
    \begin{subfigure}{.13\linewidth}
    \centering
    \includegraphics[width = \linewidth]{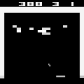}
  \end{subfigure}%
  \hspace{0.5em}%
  \begin{subfigure}{.13\linewidth}
    \centering
    \includegraphics[width = \linewidth]{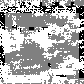}
  \end{subfigure}%
  \hspace{0.5em}%
   \begin{subfigure}{.13\linewidth}
    \centering
    \includegraphics[width = \linewidth]{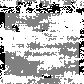}
  \end{subfigure}%
  \hspace{0.5em}%
  \begin{subfigure}{.13\linewidth}
    \centering
    \includegraphics[width = \linewidth]{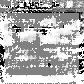}
  \end{subfigure}
  \hspace{0.5em}%
  \begin{subfigure}{.13\linewidth}
    \centering
    \includegraphics[width = \linewidth]{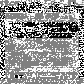}
  \end{subfigure}%
  \hspace{0.5em}%
  \begin{subfigure}{.13\linewidth}
    \centering
    \includegraphics[width = \linewidth]{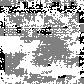}
  \end{subfigure}
  
  \begin{subfigure}{.13\linewidth}
    \centering
    \includegraphics[width = \linewidth]{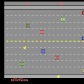}
    \caption{Clean (RGB)}
  \end{subfigure}%
  \hspace{0.5em}%
    \begin{subfigure}{.13\linewidth}
    \centering
    \includegraphics[width = \linewidth]{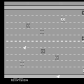}
    \caption{Clean (Gray)}
      \end{subfigure}%
  \hspace{0.5em}%
  \begin{subfigure}{.13\linewidth}
    \centering
    \includegraphics[width = \linewidth]{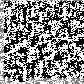}
    \caption{FGSM}
  \end{subfigure}%
  \hspace{0.5em}%
  \begin{subfigure}{.13\linewidth}
    \centering
    \includegraphics[width = \linewidth]{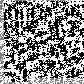}
    \caption{\osfw}
  \end{subfigure}%
  \hspace{0.5em}%
  \begin{subfigure}{.13\linewidth}
    \centering
    \includegraphics[width = \linewidth]{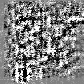}
    \caption{\uaps}
  \end{subfigure}
  \hspace{0.5em}%
  \begin{subfigure}{.13\linewidth}
    \centering
    \includegraphics[width = \linewidth]{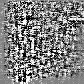}
    \caption{\uapo}
  \end{subfigure}%
  \hspace{0.5em}%
  \begin{subfigure}{.13\linewidth}
    \centering
    \includegraphics[width = \linewidth]{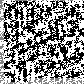}
    \caption{\osfwu}
  \end{subfigure}
  \caption{Comparison of perturbation size added into the same clean (in both RGB and gray-scale) observation in different attacks when $\epsilon=0.01$. \uaps and \uapo generate smaller perturbations. Top row: DQN agent playing Pong, Middle row: PPO agent playing Breakout, Bottom row: A2C agent playing Freeway. In perturbation images, black pixels: $-0.01$, white pixels: $+0.01$, gray pixels: $0.0$.}\label{fig:distortion}
\end{figure*}

\begin{figure*}[t]
  \centering
  \Description[<short description>]{<long description>}
  \begin{subfigure}{.13\linewidth}
    \centering
    \includegraphics[width = \linewidth]{figures/distortions/Pong_clean_rgb.png}
    \end{subfigure}%
  \hspace{0.5em}%
  \begin{subfigure}{.13\linewidth}
    \centering
    \includegraphics[width = \linewidth]{figures/distortions/Pong_clean.png}
  \end{subfigure}%
  \hspace{0.5em}%
  \begin{subfigure}{.13\linewidth}
    \centering
    \includegraphics[width = \linewidth]{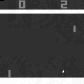}
  \end{subfigure}%
  \hspace{0.5em}%
  \begin{subfigure}{.13\linewidth}
    \centering
    \includegraphics[width = \linewidth]{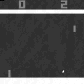}
  \end{subfigure}%
  \hspace{0.5em}%
  \begin{subfigure}{.13\linewidth}
    \centering
    \includegraphics[width = \linewidth]{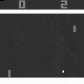}
  \end{subfigure}
  \hspace{0.5em}%
  \begin{subfigure}{.13\linewidth}
    \centering
    \includegraphics[width = \linewidth]{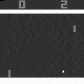}
  \end{subfigure}%
  \hspace{0.5em}%
  \begin{subfigure}{.13\linewidth}
    \centering
    \includegraphics[width = \linewidth]{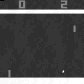}
  \end{subfigure}
  
    \begin{subfigure}{.13\linewidth}
    \centering
    \includegraphics[width = \linewidth]{figures/distortions/Breakout_clean_rgb.png}
  \end{subfigure}%
    \hspace{0.5em}%
  \begin{subfigure}{.13\linewidth}
    \centering
    \includegraphics[width = \linewidth]{figures/distortions/Breakout_clean.png}
  \end{subfigure}%
  \hspace{0.5em}%
  \begin{subfigure}{.13\linewidth}
    \centering
    \includegraphics[width = \linewidth]{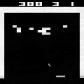}
  \end{subfigure}%
  \hspace{0.5em}%
   \begin{subfigure}{.13\linewidth}
    \centering
    \includegraphics[width = \linewidth]{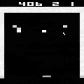}
  \end{subfigure}%
  \hspace{0.5em}%
  \begin{subfigure}{.13\linewidth}
    \centering
    \includegraphics[width = \linewidth]{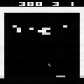}
  \end{subfigure}
  \hspace{0.5em}%
  \begin{subfigure}{.13\linewidth}
    \centering
    \includegraphics[width = \linewidth]{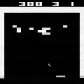}
  \end{subfigure}%
  \hspace{0.5em}%
  \begin{subfigure}{.13\linewidth}
    \centering
    \includegraphics[width = \linewidth]{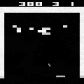}
  \end{subfigure}
  
    \begin{subfigure}{.13\linewidth}
    \centering
    \includegraphics[width = \linewidth]{figures/distortions/Freeway_clean_rgb.png}
    \caption{Clean (RGB)}
  \end{subfigure}%
    \hspace{0.5em}%
  \begin{subfigure}{.13\linewidth}
    \centering
    \includegraphics[width = \linewidth]{figures/distortions/Freeway_clean.png}
    \caption{Clean (Gray)}
  \end{subfigure}%
  \hspace{0.5em}%
  \begin{subfigure}{.13\linewidth}
    \centering
    \includegraphics[width = \linewidth]{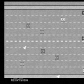}
    \caption{FGSM}
  \end{subfigure}%
  \hspace{0.5em}%
   \begin{subfigure}{.13\linewidth}
    \centering
    \includegraphics[width = \linewidth]{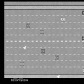}
    \caption{\osfw}
  \end{subfigure}%
  \hspace{0.5em}%
  \begin{subfigure}{.13\linewidth}
    \centering
    \includegraphics[width = \linewidth]{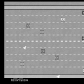}
    \caption{\uaps}
  \end{subfigure}
  \hspace{0.5em}%
  \begin{subfigure}{.13\linewidth}
    \centering
    \includegraphics[width = \linewidth]{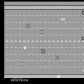}
    \caption{\uapo}
  \end{subfigure}%
  \hspace{0.5em}%
  \begin{subfigure}{.13\linewidth}
    \centering
    \includegraphics[width = \linewidth]{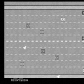}
    \caption{\osfwu}
  \end{subfigure}
  \caption{Comparison of the clean (in both RGB and gray-scale) and perturbed observations in different attacks when $\epsilon=0.01$. \uaps and \uapo generate smaller perturbations that are less visible. Top row: DQN agent playing Pong, Middle row: PPO agent playing Breakout, Bottom row: A2C agent playing Freeway.}\label{fig:distortion2}
\end{figure*}

On top of comparing the perturbation size in Pong (Figure~\ref{fig:distortionPong}, Section~\ref{ssec:attackperformance}), we also include results comparing the perturbation size for other games. As shown in Figure~\ref{fig:distortion}, FGSM generates a perturbation that contains grey patches where the sign of the gradient of the loss is zero
against the DQN agent. It has also completely black and white pixels due to the $l_{\infty}$ constraint, which makes the perturbation more visible. In addition, Figure~\ref{fig:distortion2} compares the clean and perturbed frames generated by different attacks. Note that we inject the perturbation into pre-processed frames (i.e., resized and gray-scale frames), so we show both RGB and gray-scale clean frames in Figure~\ref{fig:distortion} and Figure~\ref{fig:distortion2}. In Pong, the perturbation size in \uaps is the smallest when compared to other attacks.

\section{Optimal Parameters for \detector}
\label{sec:additionalexp_appendix}

\begin{table}[t]
	\resizebox{1.0\columnwidth}{!}{\begin{tabular}{c|c|c|ccc}
			\hline
			\bf Game & \bf  Agent                 & \bf  Parameters                                                                                                               & \bf  Attack             & \bf  Precision/Recall \\
			\hline
			\multirow{15}{*}{Pong} & \multirow{5}{*}{DQN} & \multirow{5}{*}{\begin{tabular}[c]{@{}l@{}}$k_1=12$, $k_2=24$,\\ $p=100$,\\ $r=0.9$,\\ $t_1=400$,\\ $t_2=200$\end{tabular}}  & FGSM               & 1.0/1.0  \\
			&                       &          & \osfwu & 1.0/1.0  \\
			&                       &         & \uaps              & 1.0/1.0  \\
			&                       &                                                                                                                                  & \uapo             &  1.0/1.0  \\
			&                       &                                                                                                                                  & \osfw        & 1.0/1.0  \\
			\cline{2-5}
			& \multirow{5}{*}{A2C} & \multirow{5}{*}{\begin{tabular}[c]{@{}l@{}}$k_1=12$, $k_2=24$,\\ $p=100$,\\ $r=0.9$,\\ $t_1=400$,\\ $t_2=200$\end{tabular}}  & FGSM               & 1.0/1.0  \\
			&                       &         & \osfw        &  1.0/1.0  \\
			&                       &         & \uaps              & 1.0/1.0  \\
			&                       &         & \uapo             &  1.0/1.0  \\
			&                       &         & \osfwu & 1.0/1.0  \\
			\cline{2-5}
			& \multirow{5}{*}{PPO} & \multirow{5}{*}{\begin{tabular}[c]{@{}l@{}}$k_1=12$, $k_2=24$,\\ $p=100$,\\ $r=0.9$,\\ $t_1=400$,\\ $t_2=200$\end{tabular}} & FGSM               &  1.0/1.0  \\
			&                       &         & \osfw        & 1.0/1.0  \\
			&                       &         & \uaps              & 1.0/1.0  \\
			&                       &         & \uapo             & 1.0/1.0  \\
			&                       &         & \osfwu & 1.0/1.0  \\
			\hline
			\multirow{15}{*}{Freeway} & \multirow{5}{*}{DQN} & \multirow{5}{*}{\begin{tabular}[c]{@{}l@{}}$k_1=12$, $k_2=24$,\\ $p=95$ ,\\ $r=0.8$,\\ $t_1=400$,\\ $t_2=200$\end{tabular}}  & FGSM               &  1.0/0.8  \\
			&                       &         & \osfw        &  1.0/1.0  \\
			&                       &         & \uaps              &  1.0/1.0  \\
			&                       &         & \uapo             &  1.0/1.0  \\
			&                       &         & \osfwu & 1.0/1.0  \\
			\cline{2-5}
			& \multirow{5}{*}{a2c} & \multirow{5}{*}{\begin{tabular}[c]{@{}l@{}}$k_1=12$, $k_2=24$,\\ $p=100$,\\ $r=1.0$,\\ $t_1=400$,\\ $t_2=200$\end{tabular}}  & FGSM               & 1.0/1.0  \\
			&                       &         & \osfw        &  1.0/1.0  \\
			&                       &         & \uaps              &  1.0/1.0  \\
			&                       &         & \uapo             &  1.0/1.0  \\
			&                       &         & \osfwu & 1.0/1.0  \\
			\cline{2-5}
			& \multirow{5}{*}{PPO} & \multirow{5}{*}{\begin{tabular}[c]{@{}l@{}}$k_1=12$, $k_2=24$,\\ $p=90$,\\ $r=0.9$,\\ $t_1=200$,\\ $t_2=200$\end{tabular}} & FGSM               &  1.0/1.0  \\
			&                       &         & \osfw        &  1.0/0.5  \\
			&                       &         & \uaps              & 1.0/1.0  \\
			&                       &         & \uapo             & 1.0/1.0  \\
			&                       &         & \osfwu &  1.0/0.6  \\
			\hline
			\multirow{15}{*}{Breakout} & \multirow{5}{*}{DQN} & \multirow{5}{*}{\begin{tabular}[c]{@{}l@{}}$k_1=12$, $k_2=24$,\\ $p=97$,\\ $r=0.6$,\\ $t_1=60$,\\ $t_2=40$\end{tabular}}  & FGSM               & 0.7/1.0  \\
			&                       &         & \osfw        & 0.7/1.0  \\
			&                       &         & \uaps              & 0.7/1.0   \\
			&                       &         & \uapo              & 0.7/1.0   \\
			&                       &         & \osfwu &  0.2/0.2  \\
			\cline{2-5}
			& \multirow{5}{*}{A2C} & \multirow{5}{*}{\begin{tabular}[c]{@{}l@{}}$k_1=12$, $k_2=24$,\\ $p=100$,\\ $r=0.8$,\\ $t_1=60$,\\ $t_2=40$\end{tabular}} & FGSM               & 1.0/1.0  \\
			&                       &         & \osfw        & 1.0/1.0  \\
			&                       &         & \uaps              &  1.0/0.8  \\
			&                       &         & \uapo             &  1.0/1.0  \\
			&                       &         & \osfwu & 1.0/1.0  \\
			\cline{2-5}
			& \multirow{5}{*}{PPO} & \multirow{5}{*}{\begin{tabular}[c]{@{}l@{}}$k_1=12$, $k_2=24$,\\ $p=97$,\\ $r=0.8$,\\ $t_1=60$,\\ $t_2=40$\end{tabular}}  & FGSM               & 1.0/1.0  \\
			&                       &         & \osfw        & 0.8/1.0  \\
			&                       &         & \uaps              & 0.8/1.0  \\
			&                       &         & \uapo             & 0.8/1.0  \\
			&                       &          & \osfwu & 0.8/1.0  \\
			\hline
		\end{tabular}}
		\caption{Optimal values of \detector
			to detect five different attacks: FGSM, \osfw, \uaps, \uapo, and \osfwu at $\epsilon=0.01$ for all games and agents over ten episodes.}\label{tab:fulldetection}
		\vspace{-0.5cm}
	\end{table}	
We performed a grid search to find optimal parameters for \detector in Breakout, Pong, and Freeway using three different agents (DQN, A2C, and PPO) and choose the parameters with the best F1-score. 
The optimal parameter values can be found in Table~\ref{tab:fulldetection}. The precision and recall scores of \detector against five different attacks at $\epsilon=0.01$ can also be found in this table. \detector is able to detect the presence of the adversary \adversary with perfect precision and recall for most environments and agents. However, there are some combinations of environment and agent, where \detector has a low recall and/or precision.

Attacks such as \osfwu in Freeway against PPO agents have a low action change rate (as low as $20\% - 30\%$ in some episodes), and the resulting average return is still high. This means that \adversary does not change many actions in an episode. This low action change rate is consistent with all of the combinations of the game, agent, and attack that have a low recall score.
As \detector relies on modeling the distribution of the victim \victim's action distribution during an episode, it cannot detect the presence of \adversary if there are not enough actions influenced by the attack in an episode. This means that a low action change rate from \adversary would make it difficult for \detector to detect.
The precision of \detector in Breakout is lower than other environments, since episodes in Breakout terminate very quickly (as fast as in $112$ time steps) when \victim is under attack. This means that to detect the presence of \adversary, \detector has to raise an alarm early in Breakout. 
However, \detector relies on modeling 
CAPD of the current episode, and it takes time to converge. Before CAPD converges, the anomaly score is going to be high for any episode, including episodes with no attack presence.
As an episode can terminate quickly in Breakout, \detector cannot store enough actions for CAPD to converge, and this is reflected in the reported low precision values.

\end{document}